\pdfoutput=1

\documentclass[11pt]{article}

\usepackage{ACL2023}
\usepackage{multirow} 

\usepackage[subpreambles=true]{standalone}
\usepackage{import}

\usepackage{times}
\usepackage{latexsym}
\usepackage{graphicx}
\usepackage{booktabs}
\usepackage{longtable}
\usepackage{lscape}
\usepackage{comment}
\usepackage{float}
\usepackage{booktabs}
\usepackage{graphicx}
\usepackage[normalem]{ulem}
\usepackage{bussproofs}
\usepackage{tcolorbox}
\usepackage{caption}
\usepackage{subcaption}
\usepackage{booktabs}
\usepackage{tabularx}
\usepackage{multirow}
\usepackage{graphicx}

\usepackage{xcolor,colortbl}
\usepackage{pbox}

\usepackage[T1]{fontenc}

\usepackage[utf8]{inputenc}

\usepackage{microtype}

\usepackage{inconsolata}

\title{FACTIFY-5WQA: 5W Aspect-based Fact Verification through Question Answering}

\author{\textbf{Anku Rani}$^{1}$ \quad \textbf{S.M Towhidul Islam Tonmoy}$^{2}$ \quad \textbf{Dwip Dalal}$^{3}$ \\ \textbf{Shreya Gautam}$^{4}$ \quad \textbf{Megha Chakraborty}$^{1}$ \\ 
\textbf{Aman Chadha}\dag\textsuperscript{5,6} \quad \textbf{Amit Sheth}$^{1}$ \quad \textbf{Amitava Das}$^{1}$ \quad \\
$^{1}$University of South Carolina, USA \quad   
$^{2}$IUT, Bangladesh \quad   
$^{3}$IIT Gandhinagar, India \quad \\
$^{4}$BIT Mesra, India \quad    
$^{5}$Stanford University, USA \quad  
$^{6}$Amazon AI, USA \\
\tt  arani@mailbox.sc.edu \quad
\tt amitava@mailbox.sc.edu
}

\begin{document}
\maketitle
\renewcommand{\thefootnote}{\fnsymbol{footnote}}
\footnotetext[2]{Work does not relate to the position at Amazon.}
\renewcommand*{\thefootnote}{\arabic{footnote}}
\setcounter{footnote}{0}
\begin{abstract}

Automatic fact verification has received significant attention recently. Contemporary automatic fact-checking systems focus on estimating truthfulness using numerical scores which are not human-interpretable. A human fact-checker generally follows several logical steps to verify a verisimilitude claim and conclude whether it's truthful or a mere masquerade. Popular fact-checking websites follow a common structure for fact categorization such as \textit{half true, half false, false, pants on fire}, etc. Therefore, it is necessary to have an aspect-based (\textit{delineating which part(s) are true and which are false}) explainable system that can assist human fact-checkers in asking relevant questions related to a fact, which can then be validated separately to reach a final verdict. In this paper, we propose a 5W framework (\textit{who, what, when, where, and why}) for question-answer-based fact explainability. To that end, we present a semi-automatically generated dataset called FACTIFY-5WQA, which consists of $391,041$ facts along with relevant 5W QAs -- underscoring our major contribution to this paper. A semantic role labeling system has been utilized to locate 5Ws, which generates QA pairs for claims using a masked language model. Finally, we report a baseline QA system to automatically locate those answers from evidence documents, which can serve as a baseline for future research in the field. Lastly, we propose a robust fact verification system that takes paraphrased claims and automatically validates them. The dataset and the baseline model are available at \url{https://github.com/ankuranii/acl-5W-QA}

\end{abstract}

\vspace{-4mm}
\section{Fact checking demands aspect-based explainability}
\begin{table*}[!ht]
\centering
\resizebox{0.85\textwidth}{!}{%
\begin{tabular}{clcccc}
\toprule
\multicolumn{6}{c}{\cellcolor[HTML]{9698ED}\textbf{Factify Question Answering at a glance}}                                                                                                                                                                                                                                                                                                                \\ \midrule
\textbf{Entailment Classes}                                     & \multicolumn{1}{c}{\textbf{Textual support}}                                                               & \textbf{No. of claims} & \textbf{\begin{tabular}[c]{@{}c@{}}No. of paraphrased \\ claims\end{tabular}} & \textbf{5WQA pairs} & \textbf{\begin{tabular}[c]{@{}c@{}}No. of evidence \\ documents\end{tabular}} \\ \hline
\cellcolor[HTML]{009901}{\color[HTML]{FFFFFF} \textbf{Support}} & \begin{tabular}[c]{@{}c@{}}Text are supporting \\ each other\\ $\sim$ similar news\end{tabular}                   & 217,856                & 992,503                                                                      & 464,766           & 217,635                                                                       \\ \hline
\cellcolor[HTML]{F8A102}{\color[HTML]{FFFFFF} \textbf{Neutral}} & \begin{tabular}[c]{@{}c@{}}Text are neither \\ supported nor refuted \\ $\sim$ may have common words\end{tabular} & 79,318                 & 365,593                                                                       & 194,635             & 45,715                                                                        \\ \hline
\cellcolor[HTML]{FE0000}{\color[HTML]{FFFFFF} \textbf{Refute}}  & \begin{tabular}[c]{@{}c@{}}Fake Claim \end{tabular}                                                                                                & 93,867                 & 383,035                                                                       & 243,904             & 93,766                                                                        \\ \hline
Total                                                           &                                                                                                            & 391,041                & 1,741,131                                                                     & 903,305           & 357,116  \\   \bottomrule                                                                 
\end{tabular}
}
\caption{A top-level view of Factify-5WQA: (i) classes and their respective textual support specifics, (ii) Number of claims,(iii) Number of paraphrased claims, (iv) 5WQA pairs, and (v) evidence documents}
\label{tab:glance}
\end{table*}

\begin{table*}[!t]
\centering
\resizebox{0.8\textwidth}{!}{%
\begin{tabular}{@{}lllll@{}}
\toprule
\multicolumn{5}{c}{\cellcolor[HTML]{68CBD0}
{
\begin{huge}
\textbf{5W QA based Explainability}
\end{huge}
}
}  
\\ \midrule

\multicolumn{5}{c}{\begin{tabular}[c]{@{}c@{}}
\\ 

\begin{Large}
    \textbf{\underline{Claim:}} Moderna’s lawsuits against Pfizer-BioNTech show COVID-19 vaccines were in the works before the pandemic started.
\end{Large}
\end{tabular}} \\ \midrule
 
\begin{LARGE}\textbf{Who claims}\end{LARGE}     & \begin{LARGE}\textbf{What claims}\end{LARGE}    & \begin{LARGE}\textbf{When claims}\end{LARGE}    & \begin{LARGE}\textbf{Where claims}\end{LARGE}   & \begin{LARGE}\textbf{Why claims}\end{LARGE}    
\\
\hline

\begin{tabular}[c]{@{}l@{}}
\parbox{5cm}{
\begin{Large}
\begin{itemize}
\item \textbf{Q1}: \textbf{\textit{Who lawsuits against whom?}}\\ \underline{Ans}: Moderna lawsuits against Pfizer-BioNTech\\
\end{itemize}
\end{Large}
}
\end{tabular} &

  \begin{tabular}[c]{@{}l@{}}
  \parbox{5cm}{
\begin{Large}
  \begin{itemize}
  \item \textbf{Q1}: \textbf{\textit{What the lawsuit shows?}}\\
  \underline{Ans:} COVID-19 vaccines were in the works before the pandemic started\\
  \end{itemize}
\end{Large}
  }
  \end{tabular} &
  
  \begin{tabular}[c]{@{}l@{}}
  \parbox{5cm}{
\begin{Large}
  \begin{itemize}
  \item \textbf{Q1}: \textbf{\textit{When the COVID-19 vaccines were in work?}}\\ \underline{Ans}: before pandemic.
  \end{itemize}
\end{Large}
  }
  \end{tabular} &
  
  \begin{tabular}[c]{@{}l@{}}
  \parbox{5cm}{
\begin{Large}
  \begin{itemize}
  \item no claim!
  \end{itemize}
\end{Large}
  }
  \end{tabular} &
  
  \begin{tabular}[c]{@{}l@{}}
  \parbox{5cm}{
\begin{Large}
  \begin{itemize}
  \item no claim!
  \end{itemize}
\end{Large}
  }
  \end{tabular}\\
  \hline

\includegraphics[width=0.065\textwidth]{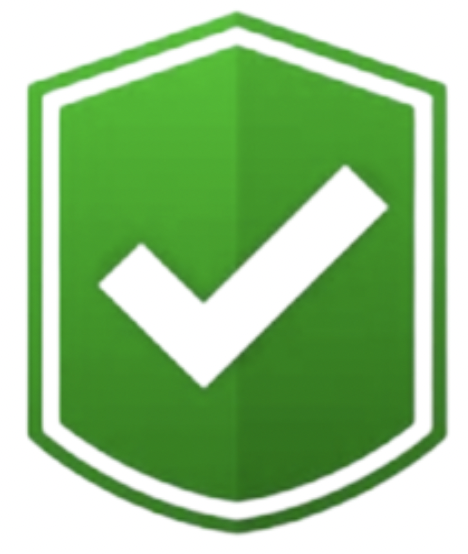}
\begin{LARGE}\textcolor{green}{\textbf{verified true}}\end{LARGE}
& 
\includegraphics[width=0.06\textwidth]{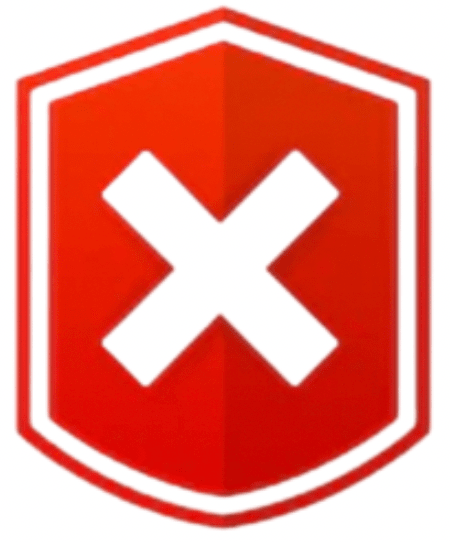} 
\begin{LARGE}\textcolor{red}{\textbf{verified false}}\end{LARGE}
& 
\includegraphics[width=0.06\textwidth]{img/false.png} 
\begin{LARGE}\textcolor{red}{\textbf{verified false}}\end{LARGE}
& 
\includegraphics[width=0.04\textwidth]{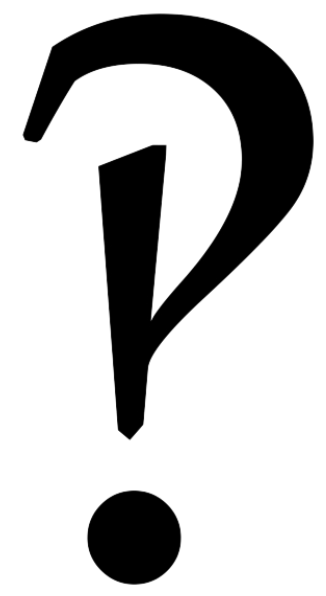}
\begin{LARGE}\textbf{not verifiable}\end{LARGE}
& 
\includegraphics[width=0.04\textwidth]{img/exclamation.png}
\begin{LARGE}\textbf{not verifiable}\end{LARGE}
\\
\hline
\multicolumn{5}{c}{\cellcolor[HTML]{C0C0C0}{
\begin{huge}
\textbf{Evidence}
\end{huge}
}
}  
\\
\hline

\begin{tabular}[c]{@{}l@{}}
\parbox{5cm}{
\begin{Large}
\begin{itemize}
\item \textbf{\textit{Moderna is suing Pfizer and BioNTech for patent infringement}}, alleging the rival companies used key parts of its mRNA technology to develop their COVID-19 vaccine. Moderna’s patents were filed between 2010 and 2016.
\end{itemize}
\end{Large}
}
\end{tabular} &

  \begin{tabular}[c]{@{}l@{}}
  \parbox{5cm}{
\begin{Large}
  \begin{itemize}
  \item Although the patents existed before the pandemic began, \textbf{\textit{this does not mean Moderna or Pfizer-BioNTech were already working on the COVID-19 vaccine}}. Scientists have used mRNA technology to study other viruses, such as the flu, Zika and rabies.
  
  \end{itemize}
\end{Large}
  }
  \end{tabular} &

  \begin{tabular}[c]{@{}l@{}}
  \parbox{5cm}{
\begin{Large}
  \begin{itemize}
  \item Although the patents existed before the pandemic began, \textbf{\textit{this does not mean Moderna or Pfizer-BioNTech were already working on the COVID-19 vaccine}}. Scientists have used mRNA technology to study other viruses, such as the flu, Zika and rabies.
\end{itemize}
\end{Large}
  }
  \end{tabular} &

  \begin{tabular}[c]{@{}l@{}}
  \parbox{5cm}{
\begin{Large}
  \begin{itemize}
\item no mention about where in any related document!
  \end{itemize}
\end{Large}
  }
  \end{tabular} &

  \begin{tabular}[c]{@{}l@{}}
  \parbox{5cm}{
\begin{Large}
  \begin{itemize}
  \item Moderna and Pfizer-BioNTech both used messenger RNA technology, or mRNA technology, to develop their COVID-19 vaccines mention where in any related document!
  \item This technology dates back to the 1990s, but the first time mRNA vaccines were widely disseminated was to combat the spread of COVID-19.
  \end{itemize}
\end{Large}
  }
  \end{tabular} 
  \\ \bottomrule
\end{tabular}
}

\caption{An illustration of 5W QA-based explainable fact verification system. This example is an illustration of the false claim. A typical semantic role labeling (SRL) system processes a sentence and identifies verb-specific semantic roles. Therefore, for the specified example, we have one sentence that has two main verbs \textit{were, and started}. For each verb, 5W QA pair will automatically be generated (2 $\times$ 5 = 10) 10 sets of QA pairs in total for this example. Further, all those $10$ 5W aspects will be fact-checked using evidence. If in case of some aspects ended having \textit{neutral} entailment verdict, possible relevant documents with associated URLs will be listed for the end user to further read and assess. This will aid human fact-checkers.}

\label{tab:5WQA_example}
\end{table*}


\vspace{-1mm}
Manual fact-checking is a time-consuming task. To assess the truthfulness of a claim, a journalist would either need to search online, offline, or both, browsing through a multitude of sources while also accounting for the perceived reliability of each source. The final verdict can then be obtained via assimilation and/or comparison of the facts derived from said sources. This process can take professional fact-checkers several hours or days \cite{10.1093/qje/qjz021} \cite{adair2017progress}, depending on the inherent complexity of the claim. 

There are several contemporary practices that journalists use for the manual verification of a claim. These methods can be categorized into four broad categories \cite{Posetti2018UNESCO}:

\vspace{-3mm}
\begin{enumerate}
    \item \textbf{Research and fact-checking}: This involves carefully researching the claim and verifying its accuracy using reliable and credible sources such as news services, academic studies, and government data. 
    \vspace{-3mm}
    \item \textbf{Interviews and expert opinions}: This involves speaking with experts in the relevant field and asking for their opinions on the claim to see if it is supported by evidence and expertise. \vspace{-3mm}
    \item \textbf{Cross-checking with multiple sources}: This involves comparing the claim with information from multiple sources to see if it is consistent or triangulates the facts obtained via multiple sources. \vspace{-3mm}
    \item \textbf{Verifying the credibility of sources}: This involves checking the credibility of the sources used to support the claim, such as ensuring that they are reliable and unbiased.
\end{enumerate}
\vspace{-3mm}

Overall, these methods can help journalists to carefully verify claims and ensure that they are accurate and supported by evidence. However, this process is tedious and hence time-consuming. A system that can generate relevant question-answer sets by dissecting the claim into its constituent components for a given verisimilitude claim could be a great catalyst in the fact-checking process.

Research on automatic fact-checking has recently received intense attention \cite{yang2022explainable},  \cite{park2021faviq},
\cite{atanasova2019automatic}, \cite{10.1162/tacl_a_00454}, \cite{trokhymovych2021wikicheck}.
Several datasets to evaluate automatic fact verification such as FEVER \cite{thorne-etal-2018-fever}, LIAR \cite{wang-2017-liar}, PolitiFact \cite{9337152}, FavIQ \cite{kwiatkowski-etal-2019-natural}, Hover \cite{jiang2020hover}, X-Fact \cite{https://doi.org/10.48550/arxiv.2106.09248}, CREAK \cite{https://doi.org/10.48550/arxiv.2109.01653}, FEVEROUS \cite{https://doi.org/10.48550/arxiv.2106.05707} are also available.  

Contemporary automatic fact-checking systems focus on estimating truthfulness using numerical scores which are not human-interpretable \cite{DBLP:journals/corr/abs-2103-07769, https://doi.org/10.48550/arxiv.2108.11896}. Others extract explicit mentions of the candidate's facts in the text as evidence for the candidate's facts, which can be hard to spot directly. Moreover, in the case of false information, it is commonplace that the whole claim isn't false, but some parts of it are, while others could still be true. A claim is either opinion-based, or knowledge-based \cite{https://doi.org/10.48550/arxiv.1804.08559}. For the same reason, the popular website Politifact based on the work by \cite{9337152} categorized the fact-checking verdict in the form of half-true, half-false, etc. 

We propose 5W (Who, What, When, Where, and Why) aspect-based question-answer pairwise explainability. Including these 5W elements within a statement can provide crucial information regarding the entities and events being discussed, thus facilitating a better understanding of the text. For instance, in the statement "\textit{Moderna's lawsuits against Pfizer-BioNTech show COVID-19 vaccines were in the works before the pandemic started.}" The use of \textit{who} highlights the individuals or entities involved in the action of filing lawsuits, \textit{what} pertains to the content of the lawsuit, specifically the revelation that COVID-19 vaccines were in the works, \textit{when} refers to the timing of this revelation, i.e., before the pandemic. Overall, the incorporation of "who," "what," "when," "where," and "why" in a text can provide crucial context and aid in making the text more clear and comprehensible.

Automatic question and answering (Q\&A) systems can provide valuable support for claims by providing evidence and supporting information. They can also help to identify potential flaws or weaknesses in a claim, allowing for further analysis and discussion. They can also help to identify potential flaws or weaknesses in a claim, allowing for further analysis and discussion.

Only two recent works \cite{9747214, kwiatkowski-etal-2019-natural} propose question answering as a proxy to fact verification explanation, breaking down automated fact-checking into several steps and providing a more detailed analysis of the decision-making processes. Question-answering-based fact explainability is indeed a very promising direction. However, open-ended QA for a fact can be hard to summarize. Therefore, we refine the QA-based explanation using the 5W framework (\textit{who, what, when, where, and why}). Journalists follow an established practice for fact-checking, verifying the so-called 5Ws \cite{10.2307/1023893}, \cite{stofer2009sports}, \cite{silverman}, \cite{su2019study}, \cite{smarts_2017}. This directs verification search and, moreover, identifies missing content in the claim that bears on its validity. One consequence of journalistic practice is that claim rejection is not a matter of degree (\textit{as conveyed by popular representations such as a number of Pinocchios or crows, or true, false, half true, half false, pants on fire}), but the rather specific, substantive explanation that recipients can themselves evaluate \cite{dobbs2012rise}.

\vspace{-1mm}
\section{Data sources and compilation}


\begin{figure*}[htbp]
    \begin{subfigure}[b]{0.33\textwidth}
    \centering
        \includegraphics[width=0.8\textwidth]{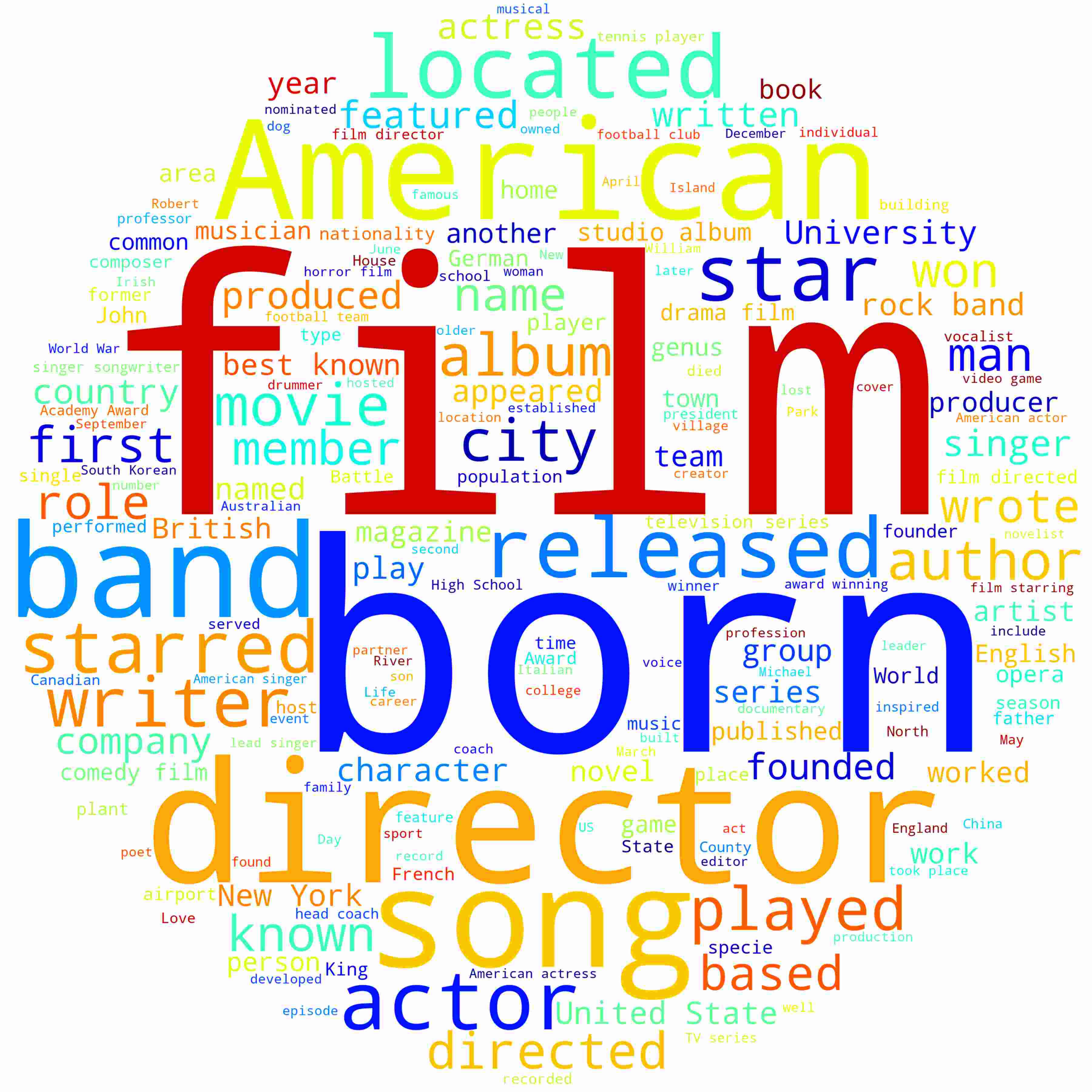}
        \caption{HoVer}
    \end{subfigure}
    \begin{subfigure}[b]{0.33\textwidth}
    \centering
        \includegraphics[width=0.8\textwidth]{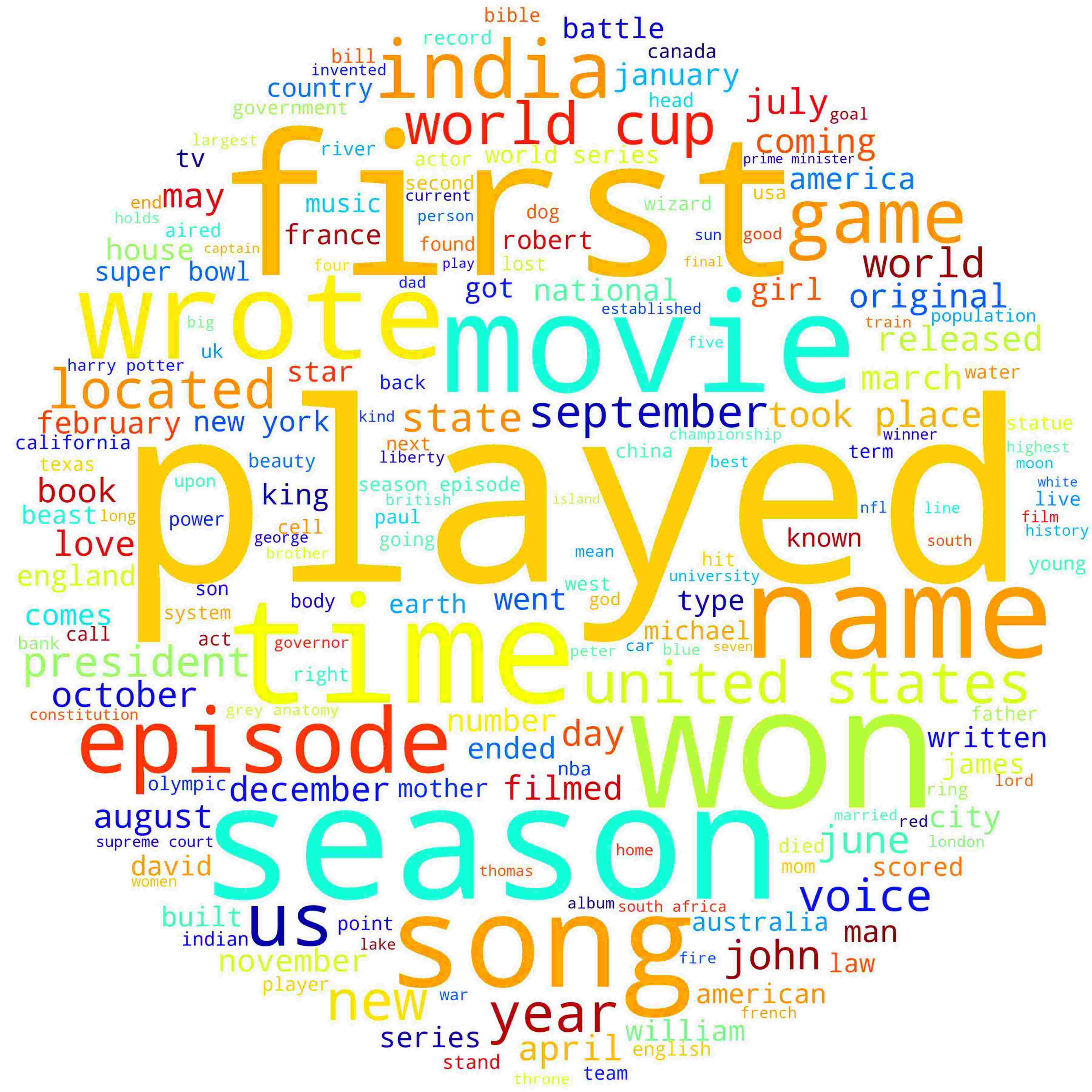}
        \caption{FaVIQ}
    \end{subfigure}
    \begin{subfigure}[b]{0.33\textwidth}
    \centering
        \includegraphics[width=0.8\textwidth]{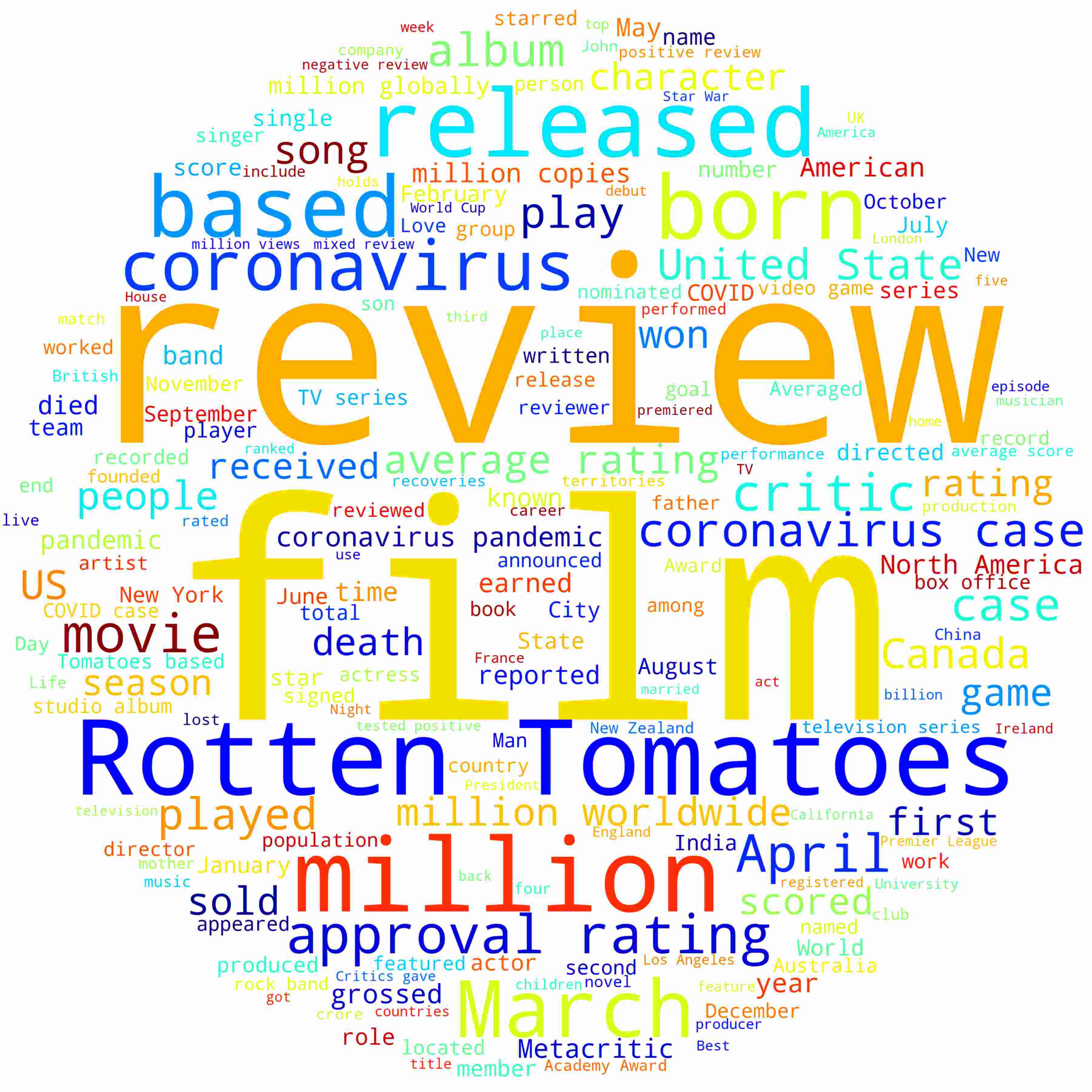}
        \caption{VITC}
    \end{subfigure}
                
    \begin{subfigure}[b]{0.33\textwidth}
    \centering
        \includegraphics[width=0.8\textwidth]{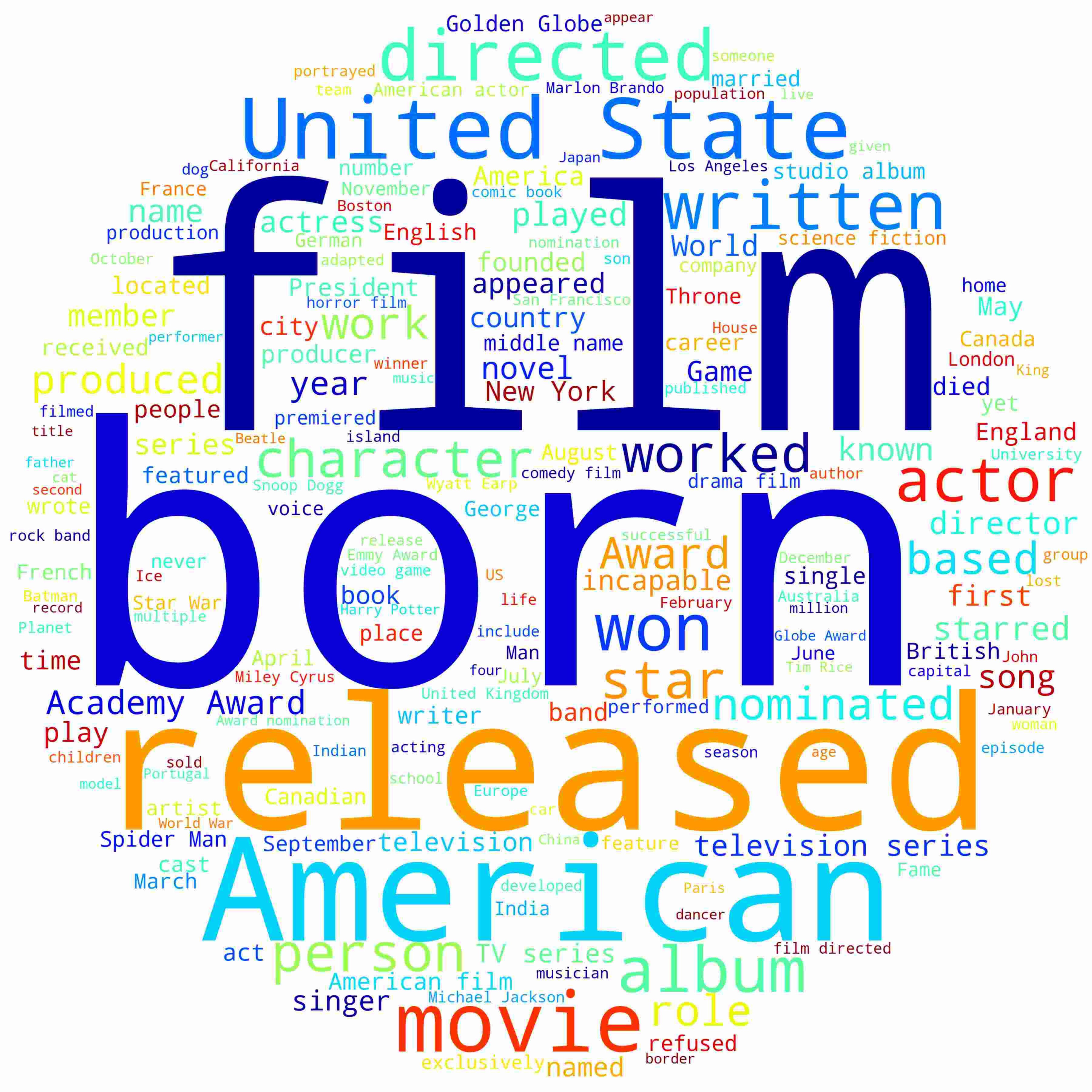}
        \caption{FEVER}
    \end{subfigure}
    \begin{subfigure}[b]{0.33\textwidth}
    \centering
        \includegraphics[width=0.8\textwidth]{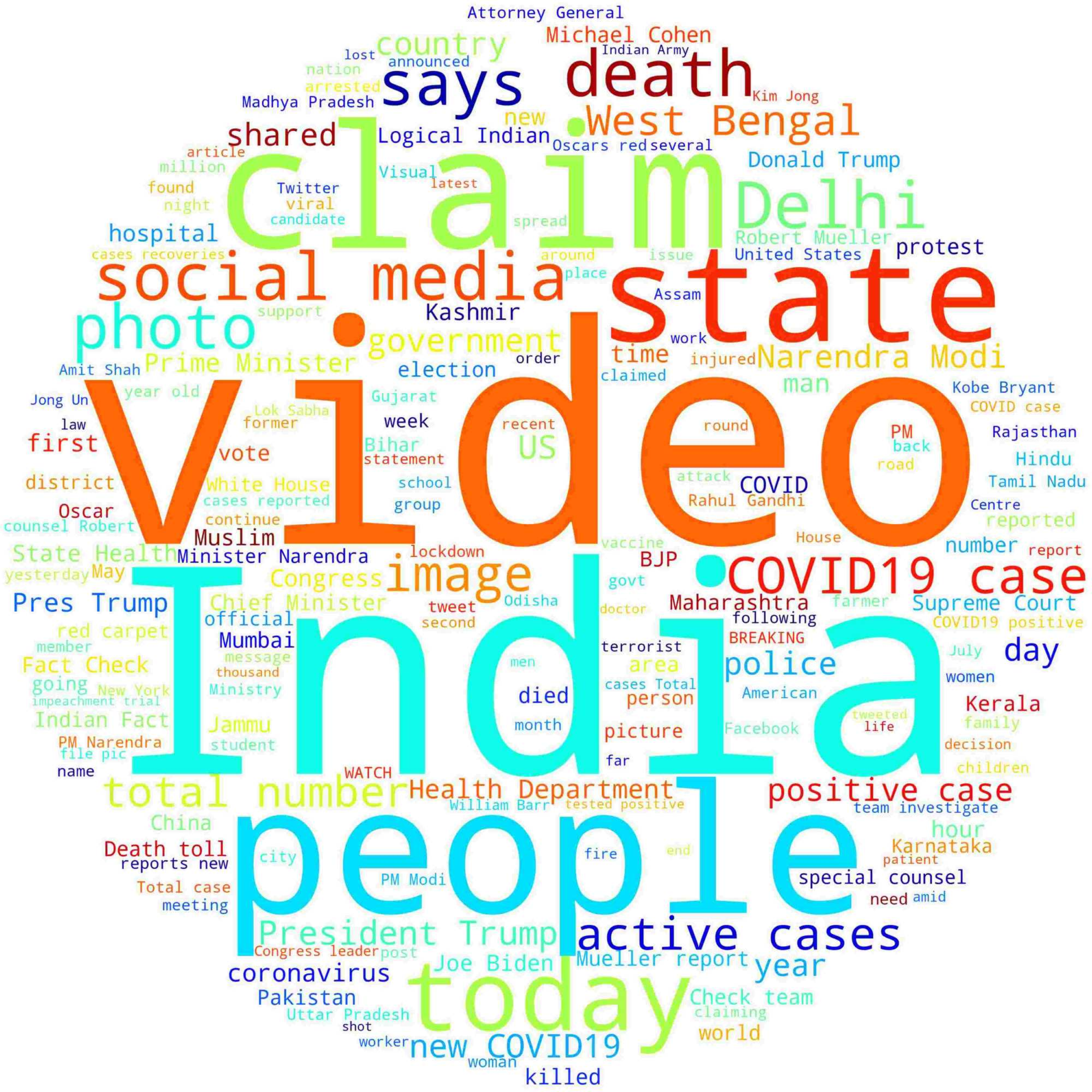}
        \caption{Factify 1.0}
    \end{subfigure}
    \begin{subfigure}[b]{0.33\textwidth}
    \centering
    \includegraphics[width=0.8\textwidth]{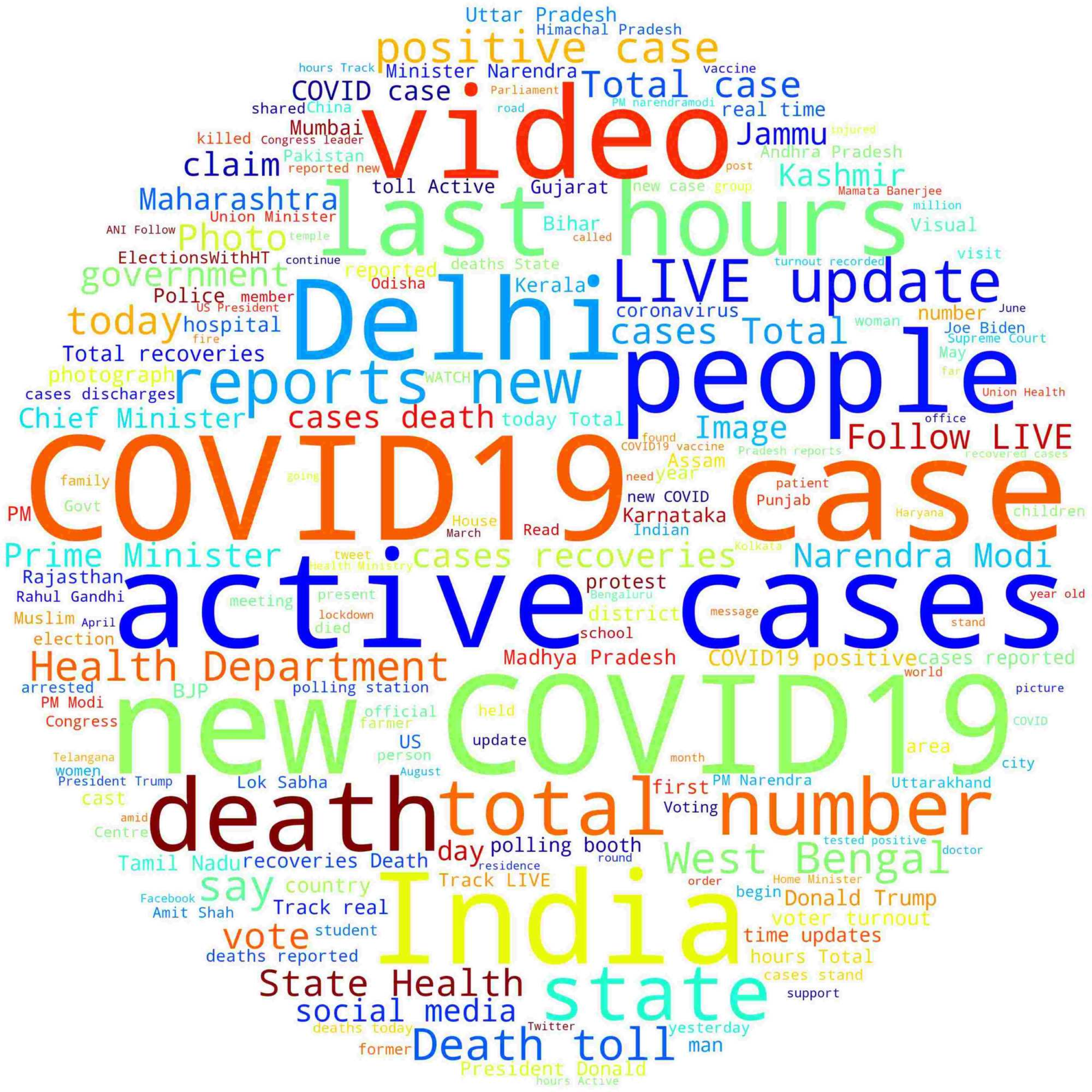}
        \caption{Factify 2.0}
    \end{subfigure}
    
    \caption{Word cloud offers a glance view of topic distributions over the chosen datasets: (i) VITC \cite{schuster2021get}, (ii) FEVER \cite{thorne2018fever}, (iii) Factify 1.0 \cite{patwa2021benchmarking}, (iv) Factify 2.0 \cite{mishra2022factify}, (v) HoVer \cite{jiang2020hover}, and (vi) FaVIQ \cite{park2021faviq}. Darker color shades in the cloud represent a higher frequency of the particular word in the dataset.}
    \label{fig:wordcloud_topic}
\end{figure*}
\vspace{-2mm}

Data collection is done by sorting 121 publicly available prevalent fact verification data sets based on modalities (111), languages (83), and tasks (51).

\begin{figure}[H]
\centering
\includegraphics[width=0.95\linewidth]{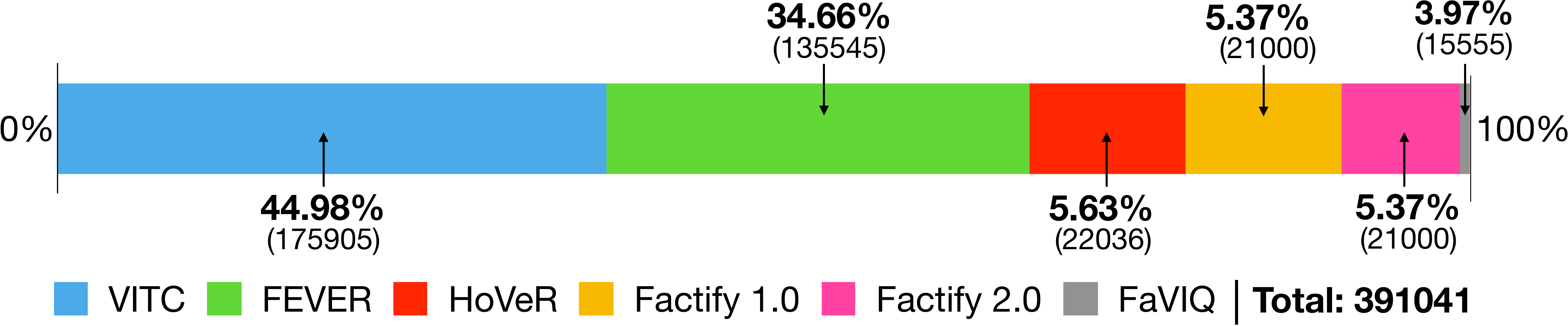}
\caption{Distribution of the FACTIFY 5WQA fact verification dataset.}
\label{fig: fact verification distribution}
\end{figure}
\vspace{-5mm}

By filtering 121 publicly available data sets for fact verification, we found ten of them to be suitable for the text-based fact verification task. We only considered the claims present in textual format in English-language because of which DanFEVER \cite{norregaard-derczynski-2021-danfever} and X-Fact \cite{https://doi.org/10.48550/arxiv.2106.09248} were also excluded because they are either Danish or multilingual. We discovered that "Evidence-based Factual Error Correction" and FEVEROUS \cite{https://doi.org/10.48550/arxiv.2106.05707} were subsets of the FEVER dataset, so we decided to use FEVER \cite{thorne2018fever}, HoVer \cite{jiang2020hover}, VITC \cite{schuster2021get}, FaVIQ \cite{park2021faviq}, Factify 1.0 \cite{patwa2021benchmarking} and Factify 2.0 \cite{mishra2022factify} for our analysis. We verified that the claims in these datasets were unique but found that 64 claims from VITC \cite{schuster2021get} overlapped with those in FEVER \cite{thorne2018fever} which is later considered once giving a total count of $391, 041$ datapoints and the distribution is represented in the figure \ref{fig: fact verification distribution}.

We only used a specific number of claims from each of the six datasets after manually inspecting the quality aspects - length of the claim and evidence, grammatical correctness, etc. For the FEVER and VITC datasets, only the claims belonging to the train split were used for making the dataset. For Factify 1.0 and Factify 2.0, the multimodal part of the dataset was discarded and only the text-based part was used. FaVIQ has two sets: the \textit{A set} and the \textit{R set}. \textit{A set} consists of ambiguous questions and their disambiguation. \textit{R set} is made by using unambiguous question-answer pairs. As discussed in earlier paragraphs, \textit{A set} is a more challenging set; hence we took the \textit{A set} of FaVIQ for making our dataset. In the case of the HoVer dataset, $22036$ claims were used in making our dataset.

We propose an amalgamated data set with the total number of unique claims as $391,041$. Around ($\sim85\%$) of them are from VITC, FEVER, and HoVer, and ($\sim15\%$) of it is from Factify 1.0, Factify 2.0  and FaVIQ as evident from Figure \ref{fig: fact verification distribution}. Figure ~\ref{fig:wordcloud_topic} offers a snapshot of topics in these datasets through a word cloud.
\vspace{-1mm}
\section{Paraphrasing textual claims}
\label{sec:section3}
\vspace{-2mm}
The motivation behind paraphrasing textual claims is as follows. A textual given claim may appear in various different textual forms in real life, owing to variations in the writing styles of different news publishing houses. Incorporating such variations is essential to developing a strong benchmark to ensure a holistic evaluation (see examples in Figure \ref{fig: paraphrase}). 

\begin{figure}[!tbh]
\centering
\resizebox{0.8\columnwidth}{!}{%
\fbox
{%
    \parbox{\columnwidth}{%
    \textcolor{blue}{Moderna’s lawsuits against Pfizer-BioNTech show COVID-19 vaccines were in the works before the pandemic started.}
    \\
    \textbf{Prphr 1:} Moderna's legal action against Pfizer-BioNTech demonstrates that work was being done on COVID-19 vaccines prior to the outbreak of the pandemic.
     \\
    \textbf{Prphr 2:} Moderna's legal action against Pfizer-BioNTech implies that work on COVID-19 vaccines had begun prior to the beginning of the pandemic.
     \\
    \textbf{Prphr 3:} Moderna's court cases against Pfizer-BioNTech indicate that COVID-19 vaccines had been in development before the pandemic began.
     \\
    \textbf{Prphr 4:} Moderna's prosecution against Pfizer-BioNTech demonstrates that COVID-19 vaccines had been in advancement prior to the pandemic commencing.   
     \\
    \textbf{Prphr 5:} It is revealed by Moderna's legal actions addressed to Pfizer-BioNTech that work on COVID-19 vaccines was being done before the pandemic began.  
    }
    }%
}
\vspace{-3mm}
\caption{Claims and paraphrases obtained using \texttt{text-davinci-003} \cite{brown2020language}}
\vspace{-4mm}
\label{fig: paraphrase}
\end{figure}
\vspace{-5mm}

 Manual generation of possible paraphrases is undoubtedly ideal, but that process is time-consuming and labor-intensive. On the other hand, automatic paraphrasing has received significant attention in recent times \cite{niu2020unsupervised} \cite{nicula2021automated}\cite{witteveen2019paraphrasing}\cite{nighojkar2021improving}. For a given claim, we generate multiple paraphrases using various SoTA models. In the process of choosing the appropriate paraphrase model based on a list of available models, the primary question we asked is how to make sure the generated paraphrases are rich in diversity while still being linguistically correct. We delineate the process followed to achieve this as follows. Let's say we have a claim $c$. We generate $n$ paraphrases using a paraphrasing model. This yields a set of $p_1^c$, $\ldots$, $p_n^c$. Next, we make pair-wise comparisons of these paraphrases with $c$, resulting in $c-p_1^c$, $\ldots$, and $c-p_n^c$. At this step, we identify the examples which are entailed, and only those are chosen. For the entailment task, we have utilized RoBERTa Large \cite{liu2019roberta} -- a SoTA model trained on the SNLI task \cite{bowman2015large}.

However, there are many other secondary factors, for e.g., a model may only be able to generate a limited number of paraphrase variations compared to others, but others can be more correct and/or consistent. As such, we considered three major dimensions in our evaluation: \textit{(i) a number of considerable paraphrase generations, (ii) correctness in those generations, and (iii) linguistic diversity in those generations}. We conducted experiments with three available models: (a) Pegasus \cite{zhang2020pegasus}, (b) T5 (T5-Large) \cite{raffel2020exploring}, and (c) GPT-3 (\texttt{text-davinci-003} variant) \cite{brown2020language}. Based on empirical observations, we concluded that GPT-3 outperformed all the other models. To offer transparency around our experiment process, we detail the aforementioned evaluation dimensions as follows.

\vspace{-2mm}
\begin{table}[H]
\centering
\resizebox{0.7\columnwidth}{!}{%
\begin{tabular}{@{}lcccc@{}}
\toprule
\textbf{Model}           &  \textbf{Coverage}  & \textbf{Correctness} & \textbf{Diversity} \\ \midrule
\textbf{Pegasus}          &   32.46   &    94.38\%       &     3.76      \\
\textbf{T5}               &  30.26       &      83.84\%       &    3.17       \\
\textbf{GPT-3} &   35.51   &   88.16\%      &     7.72      \\
\bottomrule
\end{tabular}%
}
\caption{Experimental results of automatic paraphrasing models based on three factors: \textit{(i) coverage, (ii) correctness and (iii) diversity}; GPT-3 (\texttt{text-davinci-003}) can be seen as the most performant.}
\label{tab:my-table}
\end{table}
\vspace{-4mm}

\textbf{Coverage - a number of considerable paraphrase generations:} We intend to generate up to $5$ paraphrases per given claim. Given all the generated claims, we perform a minimum edit distance (MED) \cite{wagner1974string} - units are words instead of alphabets). If MED is greater than $\pm2$ for any given paraphrase candidate (for e.g., $c-p_1^c$) with the claim, then we further consider that paraphrase, otherwise discarded. We evaluated all three models based on this setup that what model is generating the maximum number of considerable paraphrases.

\textbf{Correctness - correctness in those generations:} After the first level of filtration we have performed pairwise entailment and kept only those paraphrase candidates, are marked as entailed by the \cite{liu2019roberta} (Roberta Large), SoTA trained on SNLI \cite{bowman2015large}. 

\begin{figure}[!tbh]
\centering
\includegraphics[width=0.85\columnwidth, trim={0cm 0.5cm 0cm 0cm}]{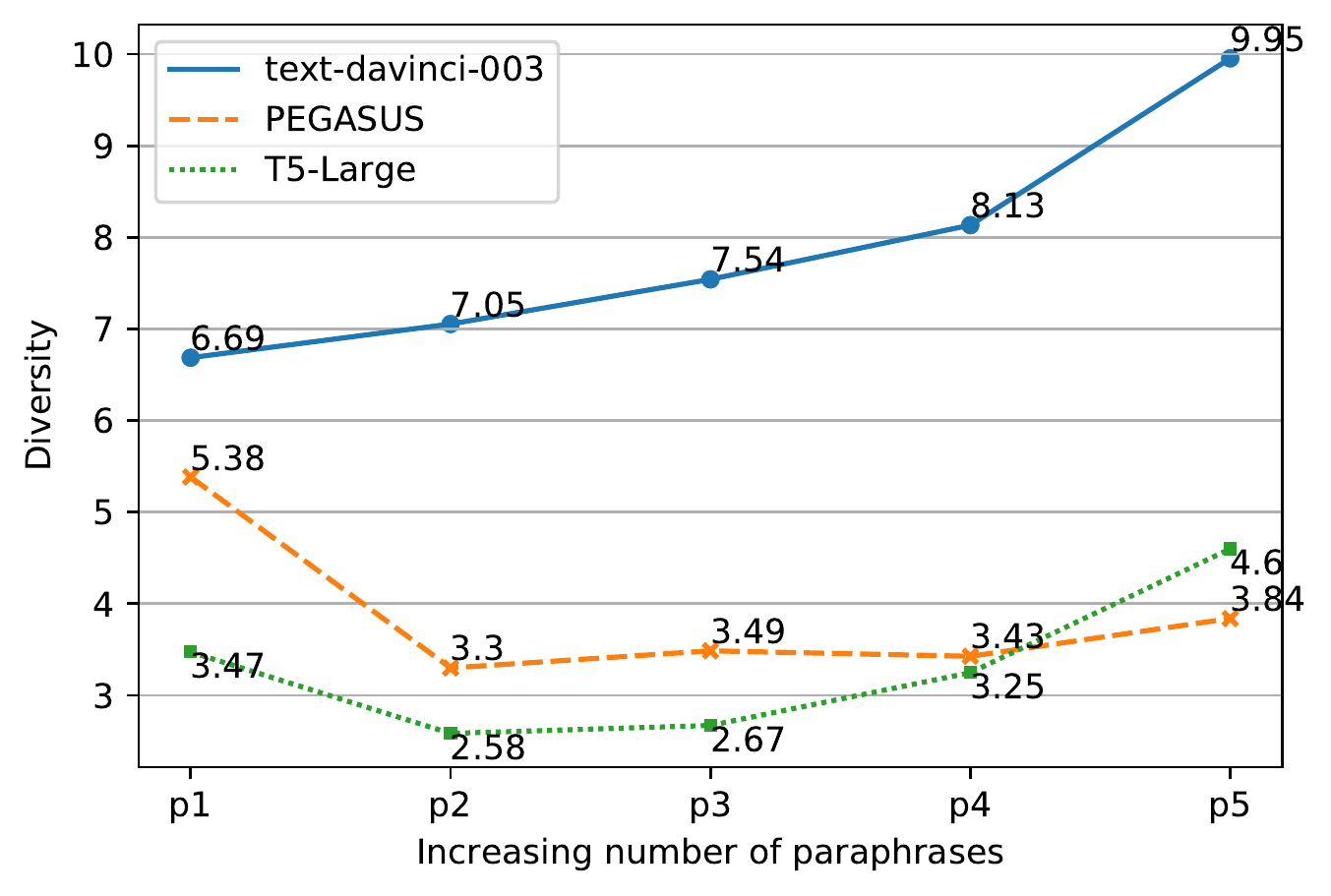}
\caption{A higher diversity score depicts an increase in the number of generated paraphrases and linguistic variations in those generated paraphrases.}
\label{fig: parr}
\end{figure}

\textbf{Diversity - linguistic diversity in those generations:} We were interested in choosing that model can produce linguistically more diverse paraphrases. Therefore we are interested in the dissimilarities check between generated paraphrase claims. For e.g., $c-p_n^c$, $p_1^c-p_n^c$, $p_2^c-p_n^c$, $\ldots$ , $p^c_{n-1}-p_n^c$ and repeat this process for all the other paraphrases and average out the dissimilarity score. There is no such metric to measure dissimilarity, therefore we use the inverse of the BLEU score \cite{papineni2002bleu}. This gives us an understanding of how linguistic diversity is produced by a given model. Based on these experiments, we found that \texttt{text-davinci-003} performed the best. The results of the experiment are reported in the following table. Furthermore, we were more interested to choose a model that can maximize the linguistic variations, and \texttt{text-davinci-003} performs on this parameter of choice as well. A plot on diversity vs. all the chosen models is reported in Figure ~\ref{fig: parr}.

\vspace{-4mm}

\vspace{-3mm}
\section{5W Semantic Role Labelling}
\vspace{-3mm}
\begin{figure}[H]
\center
\includegraphics[width=\columnwidth]{img/textimg/srl_example.tex}
\caption{Examples of the 5W semantic role labels.}
\label{fig: SRL generation example table}
\end{figure}
\vspace{-4mm}
Identification of the functional semantic roles played by various words or phrases in a given sentence is known as semantic role labelling (SRL). SRL is a well-explored area within the NLP community. There are quite a few off-the-shelf tools available: (i) Stanford SRL \cite{manning2014stanford}, (ii) AllenNLP \cite{allennlpsrl}, etc. A typical SRL system first identifies verbs in a given sentence  and then marks all the related words/phrases haven relational projection with the verb and assigns appropriate roles. Thematic roles are generally marked by standard roles defined by the Proposition Bank (generally referred to as PropBank) \cite{palmer2005proposition}, such as: \textit{Arg0, Arg1, Arg2}, and so on. We propose a mapping mechanism to map these PropBank arguments to 5W semantic roles. (look at the conversion table \ref{tab:5w-map-SRL}).

Semantic role labelling (SRL) is a natural language processing technique that involves identifying the functions of different words or phrases in a sentence. This helps to determine the meaning of the sentence by revealing the relationships between the entities in the sentence. For example, in the sentence "\textit{Moderna’s lawsuits against Pfizer-BioNTech show COVID-19 vaccines were in the works before the pandemic started,}" \textit{Moderna} would be labeled as the \textit{agent} and \textit{Pfizer-BioNTech} would be labelled as the \textit{patient}. 

\begin{table}[h]
\centering
\resizebox{0.85\columnwidth}{!}{
\begin{tabular}{ccccccc}
\toprule 
\textbf{PropBank Role }& \textbf{Who} & \textbf{What} & \textbf{When} & \textbf{Where} & \textbf{Why} & \textbf{How} \\
\midrule
\textbf{ARG0} & \textbf{84.48} & 0.00 & 3.33 & 0.00 & 0.00 & 0.00 \\
\textbf{ARG1} & 10.34 & \textbf{53.85} & 0.00 & 0.00 & 0.00 & 0.00 \\
\textbf{ARG2} & 0.00 & 9.89 & 0.00 & 0.00 & 0.00 & 0.00 \\
\textbf{ARG3} & 0.00 & 0.00 & 0.00 & 22.86 & 0.00 & 0.00 \\
\textbf{ARG4} & 0.00 & 3.29 & 0.00 & 34.29 & 0.00 & 0.00 \\
\textbf{ARGM-TMP} & 0.00 & 1.09 & \textbf{60.00} & 0.00 & 0.00 & 0.00 \\
\textbf{ARGM-LOC} & 0.00 & 1.09 & 10.00 & \textbf{25.71} & 0.00 & 0.00 \\
\textbf{ARGM-CAU} & 0.00 & 0.00 & 0.00 & 0.00 & \textbf{100.00} & 0.00 \\
\textbf{ARGM-ADV} & 0.00 & 4.39 & 20.00 & 0.00 & 0.00 & 0.06 \\
\textbf{ARGM-MNR} & 0.00 & 3.85 & 0.00 & 8.57 & 0.00 & \textbf{90.91} \\
\textbf{ARGM-MOD} & 0.00 & 4.39 & 0.00 & 0.00 & 0.00 & 0.00 \\
\textbf{ARGM-DIR} & 0.00 & 0.01 & 0.00 & 5.71 & 0.00 & 3.03 \\
\textbf{ARGM-DIS} & 0.00 & 1.65 & 0.00 & 0.00 & 0.00 & 0.00 \\
\textbf{ARGM-NEG} & 0.00 & 1.09 & 0.00 & 0.00 & 0.00 & 0.00 \\
\bottomrule
\end{tabular}
}
\caption{A mapping table from PropBank\cite{palmer2005proposition} {(\textit{Arg0, Arg1, ...})} to 5W {(\textit{who, what, when, where, and why})}.}
\label{tab:5w-map-SRL}
\end{table}

\vspace{-4mm}

\begin{table*}[!ht]
\centering
\resizebox{\textwidth}{!}{%
\begin{tabular}{@{}llccccccccccccccc@{}}
\toprule
           & \multicolumn{8}{c}{\textbf{ProphetNet}}                                  & \multicolumn{8}{c}{\textbf{BART}}                                        \\ \midrule
           & \multicolumn{4}{c}{\textbf{Claim}}     & \multicolumn{4}{c}{\textbf{+Paraphrase}} & \multicolumn{4}{c}{\textbf{Claim}}     & \multicolumn{4}{c}{\textbf{+Paraphrase}} \\
 &
  \multicolumn{1}{c}{\textbf{BLEU}} &
  \multicolumn{1}{c}{\textbf{ROUGE-L}} &
  \multicolumn{1}{c}{\textbf{Recall}} &
  \multicolumn{1}{c}{\textbf{F1}} &
  \multicolumn{1}{c}{\textbf{BLEU}} &
  \multicolumn{1}{c}{\textbf{ROUGE-L}} &
  \multicolumn{1}{c}{\textbf{Recall}} &
  \multicolumn{1}{c}{\textbf{F1}} &
  \multicolumn{1}{c}{\textbf{BLEU}} &
  \multicolumn{1}{c}{\textbf{ROUGE-L}} &
  \multicolumn{1}{c}{\textbf{Recall}} &
  \multicolumn{1}{c}{\textbf{F1}} &
  \multicolumn{1}{c}{\textbf{BLEU}} &
  \multicolumn{1}{c}{\textbf{ROUGE-L}} &
  \multicolumn{1}{c}{\textbf{Recall}} &
  \multicolumn{1}{c}{\textbf{F1}} \\
\textbf{T5-3b} &
  \cellcolor[HTML]{9AFF99}\textbf{29.22} &
  \cellcolor[HTML]{9AFF99}\textbf{48.13} &
  \cellcolor[HTML]{9AFF99}\textbf{35.66} &
  \cellcolor[HTML]{9AFF99}\textbf{38.03} &
  \cellcolor[HTML]{9AFF99}\textbf{28.13} &
  \cellcolor[HTML]{9AFF99}\textbf{46.18} &
  \cellcolor[HTML]{9AFF99}\textbf{34.15} &
  \cellcolor[HTML]{9AFF99}\textbf{36.62} &
  21.78 &
  34.53 &
  28.03 &
  28.07 &
  20.93 &
  33.57 &
  27.65 &
  27.24 \\
\textbf{T5-Large}   & 28.81 & 48.02 & 35.26 & 37.81 & 21.46  & 46.45  & 27.19 & 36.76 & 21.46 & 34.90 & 27.41 & 27.99 & 20.88  & 33.69  & 20.88 & 27.31 \\
\textbf{BERT large} & 28.65 & 46.25 & 34.55 & 36.72 & 27.27  & 44.10  & 32.95 & 35    & 20.66 & 33.19 & 25.51 & 26.44 & 19.74  & 32.34  & 25.14 & 25.71 \\ \bottomrule
\end{tabular}%
}
\caption{Selecting the best combination - 5W QAG vs. 5W QA validation.}
\label{tab: QAG-QA}
\end{table*}

The five "W"s (what, when, where, why, who) are often used to refer to the key questions that need to be answered in order to fully understand a sentence or piece of text. SRL can be seen as a way of providing answers to these questions by identifying the various roles that words and phrases play within a sentence. For example, a semantic role labeler might identify the subject of a sentence (who or what the sentence is about), the object (who or what is being acted upon), and the verb (the action being performed). In this way, semantic role labeling can be seen as a way of providing the necessary context for answering the five "W"s, and can be an important tool in natural language processing and understanding.

In this study, we use the mapping displayed in table \ref{tab:5w-map-SRL} 
and replace the roles 
that are assigned with respect to each verb as an output from SRL with 5W. According to table \ref{tab:5w-map-SRL}, it is evident that each of the 5Ws can be mapped to semantic roles. The highest percentage of mapping is taken into consideration and concluded in table \ref{tab:5w-map-SRL}.

\begin{figure}[H]
\centering
\includegraphics[width=0.85\columnwidth, trim={0.75cm 1.75cm 1cm 2cm}]{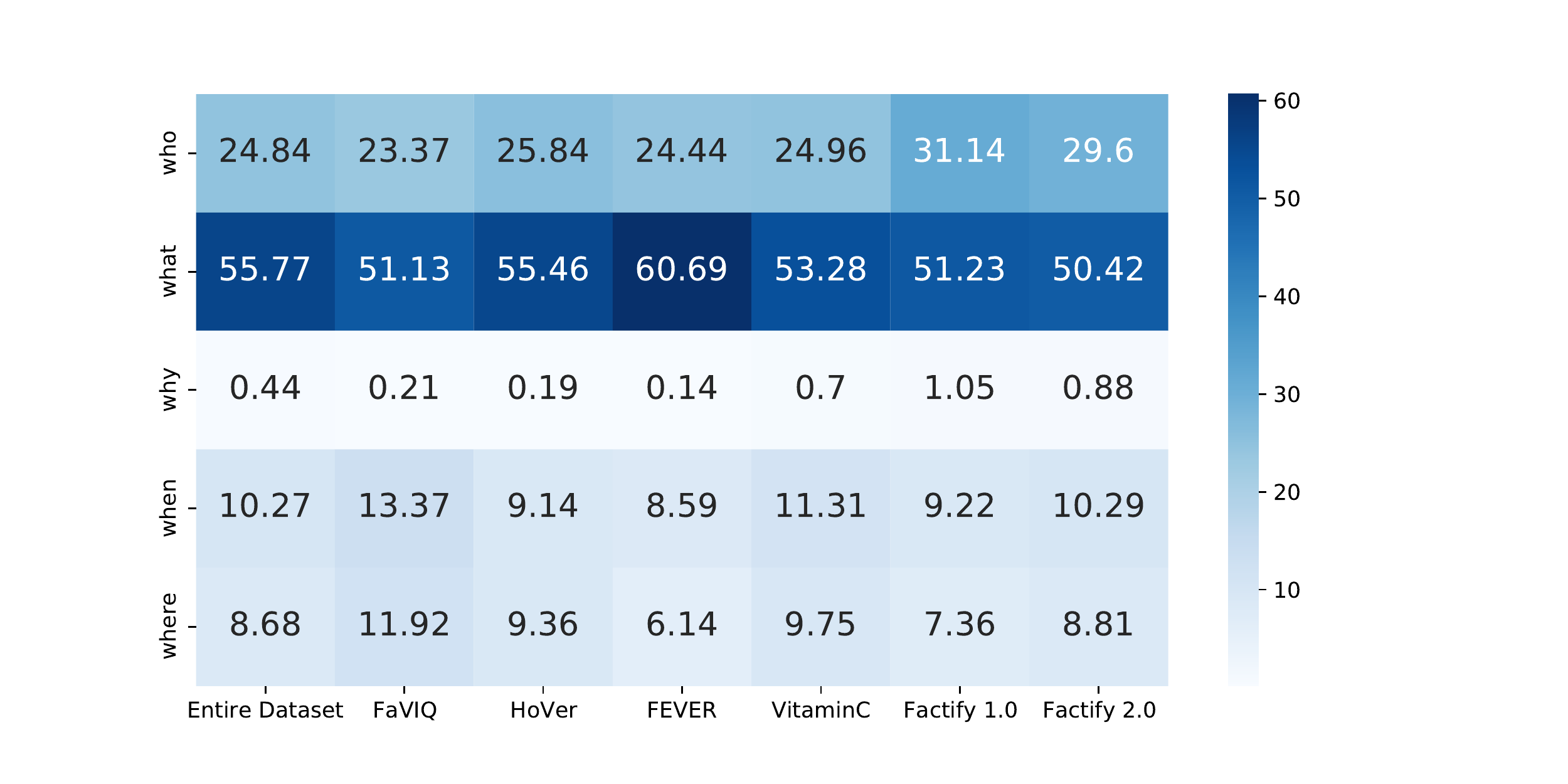}
\caption{Percentage of W's present across the dataset.}
\label{fig: nocorr}
\end{figure}

After the mapping is done, a detailed analysis for the presence of each of the 5W is conducted which is summarized in figure \ref{fig: nocorr}.

\vspace{2mm}

\vspace{-5mm}

In this study, experimentation for finding semantic roles was conducted using \textbf{AllenNLP SRL} demo \cite{allennlpsrl}. Developed by \cite{Shi2019SimpleBM}, it is a BERT \cite{devlin2018bert} based model with some modifications that introduce a linear classification layer with no additional parameters, and it is currently the best single model for English PropBank SRL on newswire sentences with a test F1 of 86.49 on the Ontonotes 5.0 dataset \cite{palmer2005proposition}.
Newswire instances correlate with the fact verification dataset as true news is also a fact. 

As indicated in figure \ref{fig: SRL generation example table}, the pipeline for generating 5W aspect-based semantic role labeling is to pass it through an SRL model and map it with 5W. An example of a claim as per the output using AllenNLP's SRL model is in figure \ref{fig: SRL generation example table}.

\vspace{-2mm}
\subsection{Human Evaluation of the 5W SRL}
\vspace{-1mm}
In this work evaluation for the 5W Aspect, based on semantic role labeling is conducted using \textit{mapping accuracy}: This involves accuracy on SRL output mapped with 5Ws.

For the purpose of finding how good the mapping of 5W is with semantic roles and generation of semantic roles, human annotation of $3000$ data points was conducted. $500$ random data points each from FEVER , FavIQ, HoVer, VITC, Factify 1.0 and Factify 2.0 were annotated and the results are described in table \ref{tab:human_annotation}.

\vspace{-3mm}
\begin{table}[H]
\centering
\resizebox{\columnwidth}{!}{
      
        \begin{tabular}{ccccccc}
\toprule
\multicolumn{1}{l}{} & \textbf{FaVIQ}               & \textbf{FEVER}                & \textbf{HoVer}                & \textbf{VitaminC}            & \textbf{Factify 1.0}            & \textbf{Factify 2.0}            \\
\hline
\textbf{Who}         & \cellcolor[HTML]{E4DC4D}89\% & \cellcolor[HTML]{FFDE50}85\%  & \cellcolor[HTML]{D6D94C}90\%  & \cellcolor[HTML]{FFE24F}87\% & \cellcolor[HTML]{FFE050}86\% & \cellcolor[HTML]{FFD950}82\% \\
\textbf{What}        & \cellcolor[HTML]{FFDE50}85\% & \cellcolor[HTML]{FFAF51}56\%  & \cellcolor[HTML]{FFC250}68\%  & \cellcolor[HTML]{FFD350}78\% & \cellcolor[HTML]{FFD850}81\% & \cellcolor[HTML]{ACCF49}93\% \\
\textbf{When}        & \cellcolor[HTML]{FFE050}86\% & \cellcolor[HTML]{D6D94C}90\%  & \cellcolor[HTML]{90C948}95\%  & \cellcolor[HTML]{66BF45}98\% & \cellcolor[HTML]{FFDB50}83\% & \cellcolor[HTML]{FFCE50}75\% \\
\textbf{Where}       & \cellcolor[HTML]{ACCF49}93\% & \cellcolor[HTML]{4AB842}100\% & \cellcolor[HTML]{D6D94C}90\%  & \cellcolor[HTML]{74C245}97\% & \cellcolor[HTML]{ACCF49}93\% & \cellcolor[HTML]{FFE050}86\% \\
\textbf{Why}         & \cellcolor[HTML]{FF5353}0\%  & -                             & \cellcolor[HTML]{4AB842}100\% & \cellcolor[HTML]{BAD24A}92\% & \cellcolor[HTML]{FFE24F}87\% & \cellcolor[HTML]{ACCF49}93\%

\\
\bottomrule
\end{tabular}
}
\caption{Human evaluation of 5W SRL; \% represents human agreement on 5W mapping with SRL.}
\label{tab:human_annotation}
\end{table}
\vspace{-5mm}

\vspace{-2mm}
\section{5W aspect-based QA pair generation}
\vspace{-2mm}
A false claim is very likely to have some truth in it, some correct information. In fact, most fake news articles are challenging to detect precisely because they are mostly based on correct information, deviating from the facts only in a few aspects. That is, the misinformation in the claim comes from a very specific inaccurate statement. So, given our textual claim, we generate 5W question-answer pairs by doing semantic role labeling on the given claim. The task is now based on the generated QA pairs, a fact-checking system can extract evidence sentences from existing authentic resources to verify or refute the claim based on each question- \textit{Who, What, When, Where, and Why} \cite{article_2023}. Please see examples in Figure ~\ref{fig: Question generation example table}.

\vspace{-3mm}
\begin{figure}[H]
\center
\includegraphics[width=\columnwidth]{img/textimg/QA_example.tex}
\caption{Examples of QA pairs generated from a claim by the QG system.}
\label{fig: Question generation example table}
\end{figure}
\vspace{-4mm}

Our method of using 5W SRL to generate QA pairs and then verify each aspect separately allows us to detect `\textit{exactly where the lie lies}'. This, in turn, provides an explanation of why a particular claim is refutable since we can identify exactly which part of the claim is false. 

\vspace{-4mm}
\begin{figure}[H]
    \centering
\includegraphics[width=\columnwidth, trim={0 0.7cm 0 0}]{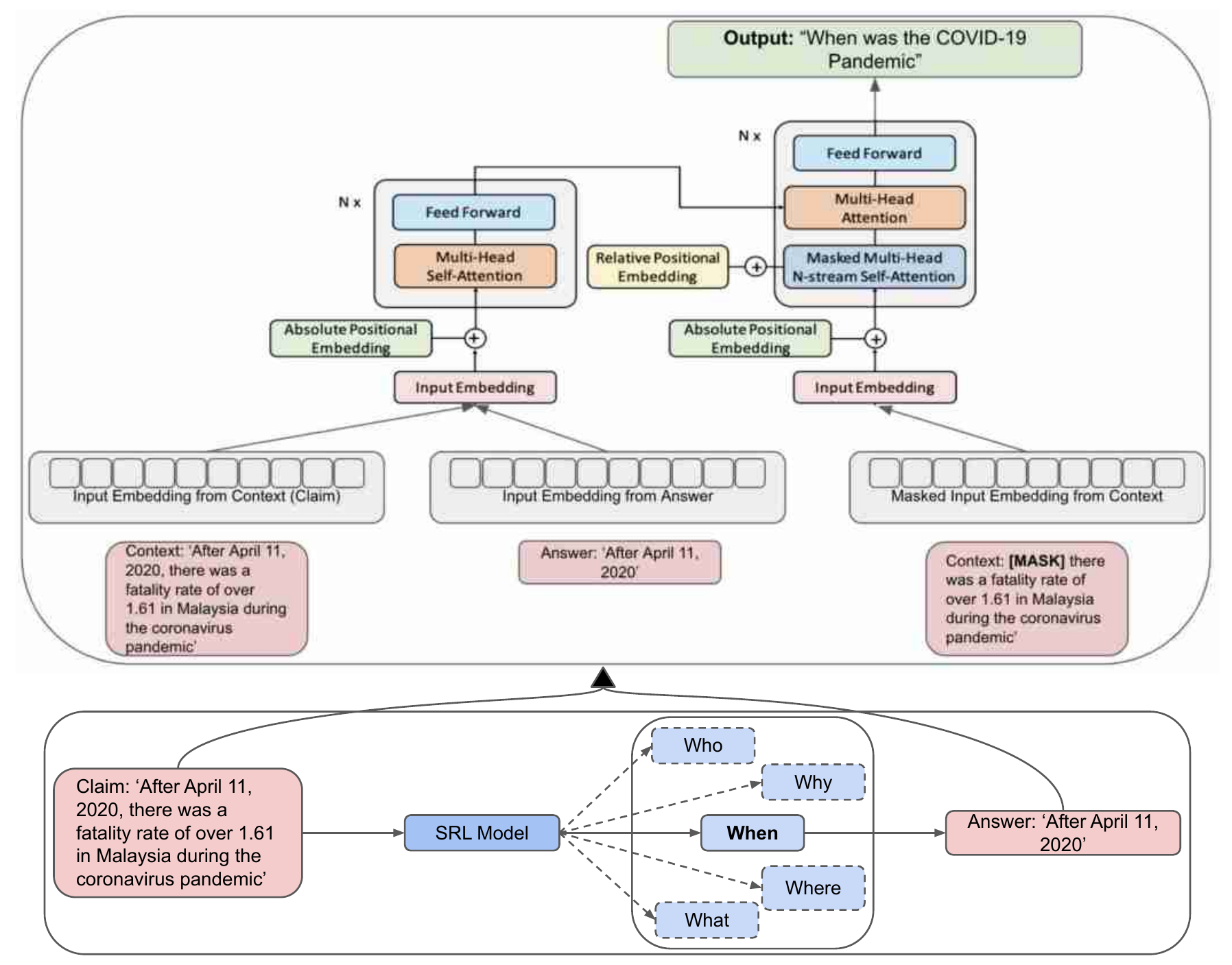}
    \caption{Illustration of 5W QA Generation Pipeline using ProphetNet.}
    \label{fig:5w_qa_gen}
\end{figure}
\vspace{-6mm}

The process of fact verification is inherently intricate, with several questions representing the components within the underlying claim that need answers to reach a verdict on the veracity of the claim. Referring to the example in figure ~\ref{fig: Question generation example table}, such questions may include: \textit{(a) Who lawsuit against whom? (b) Vaccine were in use when?}
what can go wrong if this claim is false? Manual fact-checking can be labor-intensive, consuming several hours or days \cite{10.1145/2806416.2806652, adair2017progress}.

For the 5W question generation task we have experimented with two models: (i) BART \cite{DBLP:journals/corr/abs-1910-13461}, and (ii) ProphetNet \cite{qi-etal-2020-prophetnet}, and found ProphetNet outperforms the former.

ProphetNet \cite{qi-etal-2020-prophetnet}, a generative model that uses multi-lingual pre-training with masked span generation. It is optimized through \textit{n-step} ahead prediction, which predicts the next \textit{n} tokens based on previous context tokens at each time step, encouraging the model to explicitly plan for future tokens. In this work, we employed the context-based question generation approach to generate relevant and specific questions for the task of fact verification. This approach utilizes the claim information to ensure that the generated questions are appropriate for fact-checking.


\vspace{-3mm}
\subsection{Human evaluation of QA generation}
\vspace{-1mm}

\begin{table}[H]
\centering
\resizebox{\columnwidth}{!}{

        \begin{tabular}{cccccccc}
\toprule
\multicolumn{1}{l}{}    & \multicolumn{1}{l}{}    & \multicolumn{1}{c}{\textbf{FaVIQ}}    & \multicolumn{1}{c}{\textbf{FEVER}}    & \multicolumn{1}{c}{\textbf{HoVer}}     & \multicolumn{1}{c}{\textbf{VitaminC}} & \multicolumn{1}{c}{\textbf{Factify 1.0}} & \multicolumn{1}{c}{\textbf{Factify 2.0}} \\
\hline
                        & \textbf{Question is well-formed} & \cellcolor[HTML]{FFE24F}86\% & \cellcolor[HTML]{FFD350}77\% & \cellcolor[HTML]{FFDE50}84\%  & \cellcolor[HTML]{FFD650}79\% & \cellcolor[HTML]{FFD850}80\% & \cellcolor[HTML]{FFDB50}82\% \\
                        & \textbf{Question is correct}     & \cellcolor[HTML]{D0D94F}90\% & \cellcolor[HTML]{FFDB50}82\% & \cellcolor[HTML]{FFE24F}86\%  & \cellcolor[HTML]{FFDD50}83\% & \cellcolor[HTML]{F4E04F}87\% & \cellcolor[HTML]{DCDB4F}89\% \\
\multirow{-3}{*}{\textbf{Who}}  & \textbf{Answer is correct}       & \cellcolor[HTML]{DCDB4F}89\% & \cellcolor[HTML]{FFE050}85\% & \cellcolor[HTML]{D0D94F}90\%  & \cellcolor[HTML]{F4E04F}87\% & \cellcolor[HTML]{FFE24F}86\% & \cellcolor[HTML]{FFDB50}82\% \\
\hline
                        & \textbf{Question is well-formed} & \cellcolor[HTML]{FFC950}71\% & \cellcolor[HTML]{FFAB51}53\% & \cellcolor[HTML]{FFC450}68\%  & \cellcolor[HTML]{FFD650}79\% & \cellcolor[HTML]{FFD350}77\% & \cellcolor[HTML]{FFCA50}72\% \\
                        & \textbf{Question is correct}     & \cellcolor[HTML]{FFD350}77\% & \cellcolor[HTML]{FFC550}69\% & \cellcolor[HTML]{FFC750}70\%  & \cellcolor[HTML]{FFD950}81\% & \cellcolor[HTML]{FFD850}80\% & \cellcolor[HTML]{FFD150}76\% \\
\multirow{-3}{*}{\textbf{What}}  & \textbf{Answer is correct}       & \cellcolor[HTML]{FFE050}85\% & \cellcolor[HTML]{FFB051}56\% & \cellcolor[HTML]{FFC450}68\%  & \cellcolor[HTML]{FFD450}78\% & \cellcolor[HTML]{FFD950}81\% & \cellcolor[HTML]{ACD14F}93\% \\
\hline
                        & \textbf{Question is well-formed} & \cellcolor[HTML]{E8DE4F}88\% & \cellcolor[HTML]{FFD350}77\% & \cellcolor[HTML]{FFE24F}86\%  & \cellcolor[HTML]{FFD450}78\% & \cellcolor[HTML]{FFD950}81\% & \cellcolor[HTML]{FFD450}78\% \\
                        & \textbf{Question is correct}     & \cellcolor[HTML]{D0D94F}90\% & \cellcolor[HTML]{FFE24F}86\% & \cellcolor[HTML]{E8DE4F}88\%  & \cellcolor[HTML]{A0CF4F}94\% & \cellcolor[HTML]{B8D44F}92\% & \cellcolor[HTML]{DCDB4F}89\% \\
\multirow{-3}{*}{\textbf{When}}  & \textbf{Answer is correct}       & \cellcolor[HTML]{FFE24F}86\% & \cellcolor[HTML]{D0D94F}90\% & \cellcolor[HTML]{94CD4F}95\%  & \cellcolor[HTML]{70C54F}98\% & \cellcolor[HTML]{FFDD50}83\% & \cellcolor[HTML]{FFCF50}75\% \\
\hline
                        & \textbf{Question is well-formed} & \cellcolor[HTML]{D0D94F}90\% & \cellcolor[HTML]{94CD4F}95\% & \cellcolor[HTML]{FFC450}68\%  & \cellcolor[HTML]{F4E04F}87\% & \cellcolor[HTML]{C4D64F}91\% & \cellcolor[HTML]{E8DE4F}88\% \\
                        & \textbf{Question is correct}     & \cellcolor[HTML]{FFE050}85\% & \cellcolor[HTML]{94CD4F}95\% & \cellcolor[HTML]{FFD450}78\%  & \cellcolor[HTML]{B8D44F}92\% & \cellcolor[HTML]{B8D44F}92\% & \cellcolor[HTML]{FFDD50}83\% \\
\multirow{-3}{*}{\textbf{Where}} & \textbf{Answer is correct }      & \cellcolor[HTML]{ACD14F}93\% & \cellcolor[HTML]{7CC84F}97\% & \cellcolor[HTML]{D0D94F}90\%  & \cellcolor[HTML]{7CC84F}97\% & \cellcolor[HTML]{ACD14F}93\% & \cellcolor[HTML]{FFE24F}86\% \\
\hline
                        & \textbf{Question is well-formed} & \cellcolor[HTML]{FF5353}0\%  & -                            & \cellcolor[HTML]{58C050}100\% & \cellcolor[HTML]{B8D44F}92\% & \cellcolor[HTML]{B8D44F}92\% & \cellcolor[HTML]{D0D94F}90\% \\
                        & \textbf{Question is correct}     & \cellcolor[HTML]{FF5353}0\%  & -                            & \cellcolor[HTML]{58C050}100\% & \cellcolor[HTML]{94CD4F}95\% & \cellcolor[HTML]{94CD4F}95\% & \cellcolor[HTML]{A0CF4F}94\% \\
\multirow{-3}{*}{\textbf{Why}}   & \textbf{Answer is correct}       & \cellcolor[HTML]{FF5353}0\%  & -                            & \cellcolor[HTML]{58C050}100\% & \cellcolor[HTML]{88CA4F}96\% & \cellcolor[HTML]{F4E04F}87\% & \cellcolor[HTML]{ACD14F}93\%

\\
\bottomrule
\end{tabular}
}
\caption{Human evaluation of QA generation. \% represents human agreement on how well the question is formed, and whether the question and answer are correct.}
\end{table}

For the evaluation purpose, a random sample of $3000$ data points was selected for annotation. The questions generated using the Prophetnet model were utilized for this purpose. The annotators were instructed to evaluate the question-answer pairs in three dimensions: the question is well formed, which means it is syntactically correct, the question is correct which  means it is semantically correct with respect to the given claim, and extracted answer from the model is correct. The evaluation results for the datasets are presented in the following analysis.




\vspace{-1mm}

\section{The 5W QA validation system}
\vspace{-1mm}

Finally, we propose a QA validation system, where the generated questions from the QG system and the evidence are passed through SoTA Question answering models (T5:3B) \cite{raffel2020exploring}, T5:Large \cite{raffel2020exploring}, Bert: Large \cite{devlin2018bert}) demonstrated in figure \ref{fig: Question answering example table evidence}. This helps to find out whether the evidence supports or refutes the claim or if the system misses out on enough information to make a conclusion.

\begin{figure}[H]
\center
\includegraphics[width=\columnwidth]{img/textimg/QA_pair_evidence_example1.tex}
\caption{Examples of QA pairs generated from evidence by the QA system.}
\label{fig: Question answering example table evidence}
\end{figure}
\vspace{-4mm}

\begin{figure}[H]
    \centering
\includegraphics[width=0.85\columnwidth, trim = {0.15cm 0.6cm 0 0}]{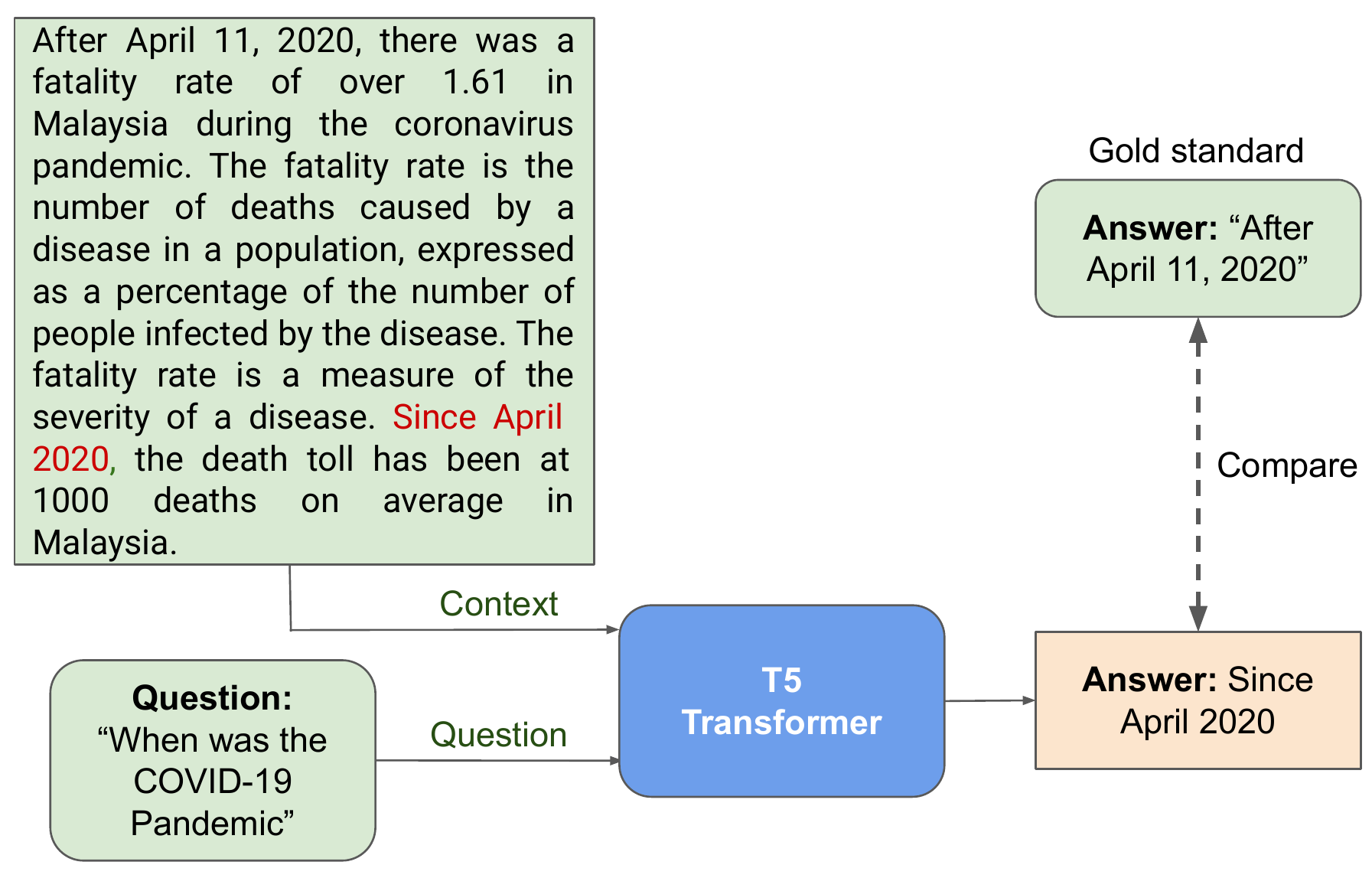}
    \caption{T5-based question answering framework.}
    \label{fig:Question answering Framework}
\end{figure}

An example of two of the claims that generate answers based on the evidence is represented in figure \ref{fig: Question answering example table evidence}. In this figure, the question is generated using prophetnet, and the answer is generated using the T5-3B model from the evidence of the claims. as described in figure \ref{fig:Question answering Framework}. 

\vspace{-3mm}
\vspace{-5mm}

To design the 5W QA validation system, we utilized the claims, evidence documents, and 5W questions generated by the question generation system as input. The answer generated by the 5W QG model is treated as the gold standard for comparison between claim and evidence. We experimented with three models, T5-3B \cite{raffel2020exploring}, T5-Large \cite{raffel2020exploring}, and Bert-Large \cite{devlin2018bert}. The T5 is an encoder-decoder-based language model that treats this task as text-to-text conversion, with multiple input sequences and produces an output as text. The model is pre-trained using the C4 corpus \cite{raffel2020exploring} and fine-tuned on a variety of tasks. T5-Large employs the same encoder-decoder architecture as T5-3B \cite{raffel2020exploring}, but with a reduced number of parameters. The third model that we experimented with is the Bert-Large \cite{devlin2018bert} model, which utilizes masked language models for pre-training, enabling it to handle various downstream tasks.


\vspace{-1mm}
\section{Selecting the best combination - 5W QAG vs. 5W QA validation}
\label{sec:section7}
\vspace{-1mm}
We have utilized off-the-self models both for 5W question-answer generation and 5W question-answer validation. Given that the datasets used for training the models bear an obvious discrepancy in terms of the distribution characteristics compared to our data (world news) which would probably lead to a generalization gap, it was essential to experimentally judge which system offered the best performance for our use-case. Instead of choosing the best system for generation vs. validation, we opted for pair-wise validation to ensure we chose the best combination. Table \ref{tab: QAG-QA} details our evaluation results -- the rows denote the QA models while the columns denote QAG models. From the results in the table, we can see that the best combination in terms of a QAG and QA validation model was identified as T5-3b and ProphetNet, respectively.
\section{Conclusion and future avenues}
It has been realized by the community that due to the given complexity of fact-checking it possibly can not be automated completely. Human-in-loop is the solution for the same. Proposed 5W QA-based fact verification can be the best aid for human fact-checkers. To the best of our knowledge, we are the first to introduce 5W QA-based fact verification and additionally proposed relevant techniques to automatically generate QA using the automatic method, which can be readily used for any incoming claim on the spot. Furthermore, the QA validation section can aid to provide evidence support. Paraphrasing claims provide a holistic approach to fact-checking. Generated datasets and resources will be made public for research purposes containing 3.91 million claims.

\section{Discussion and limitations}
In this section, we self-criticize a few aspects that could be improved and also detail how we plan (tentatively) to plan to improve upon those specific aspects - 

\subsection{Paraphrasing claims}
\label{sec:limitations_paraphrase}
Manual generation of possible paraphrases is undoubtedly ideal but is time-consuming and labor-intensive. Automatic paraphrasing is a good way to scale quickly, but there could be more complex variations of meaning paraphrases hard to generate automatically. For example - "\textit{It's all about business - a patent infringement case against Pfizer by a rival corporate reveals they knew about COVID in one way!}" and "\textit{Oh my god COVID is not enough now we have to deal with HIV blood in the name of charity!}". 

An ideal for this shortcoming would be to manually generate a few thousand paraphrase samples and then fine-tune language models. On the other hand, a new paradigm in-context Learning is gaining momentum \cite{7580601}. In-context learning has been magical in adapting a language model to new tasks through just a few demonstration examples without doing gradient descent. There are quite a few recent studies that demonstrate new abilities of language models that learn from a handful of examples in the context (in-context learning - ICL for short). Many studies have shown that LLMs can perform a series of complex tasks with ICL, such as solving mathematical reasoning problems \cite{https://doi.org/10.48550/arxiv.2201.11903}. These strong abilities have been widely verified as emerging abilities for large language models \cite{https://doi.org/10.48550/arxiv.2201.11903}. From prompt engineering to chain of thoughts, we are excited to do more experiments with the new paradigm of in-context learning for automatically paraphrasing claims.

\subsection{5W SRL}
\label{sec:limitations_5wsrl}
Semantic role labeling is a well-studied  sub-discipline, and the mapping mechanism we proposed works well in most cases except in elliptic situations like anaphora and cataphora. In the future, we would like to explore how an anaphora and coreference resolution \cite{joshi2019coref} can aid an improvement.

\subsection{5W QA pair generation}
\label{sec:limitations_qg}
5W semantic role-based question generation is one of the major contributions of this paper. While automatic generation aided in scaling up the QA pair generation, it also comes with limitations of generating more complex questions covering multiple Ws and \textit{how} kinds of questions; for example, "\textit{How Moderna is going to get benefited if this Pfizer COVID news turns out to be a rumor?}". For the betterment of FACTIFY benchmark, we would like to generate few thousand manually generated abstract QA pairs. Then will proceed 
towards in-context Learning \cite{7580601}.

Abstractive question-answering has received momentum \cite{zhao-etal-2022-compositional}, \cite{pal-etal-2022-parameter} recently. We want to explore how we can generate more abstract QA pairs for the multimodal fact-verification task.

\subsection{QA system for the 5W question}
\label{sec:limitations_qa}
Generated performance measures attest the proposed QA model needs a lot more improvement. This is due to the complexity of the problem and we believe that will attract future researchers to try this benchmark and conduct research on multimodal fact verification. 

It has been realized by the community that relevant document retrieval is the major bottleneck for fact verification. Recent work introduced a fresh perspective to the problem - named Hypothetical Document Embeddings (HyDE) \cite{hyde} and applied a clever trick even if the wrong answer is more semantically similar to the right answer than the question. This could be an interesting direction to explore and examine how that could aid in retrieving relevant documents and answers.

\bibliography{anthology,custom}

\begin{thebibliography}{57}
\expandafter\ifx\csname natexlab\endcsname\relax\def\natexlab#1{#1}\fi

\bibitem[{Adair et~al.(2017)Adair, Li, Yang, and Yu}]{adair2017progress}
Bill Adair, Chengkai Li, Jun Yang, and Cong Yu. 2017.
\newblock Progress toward “the holy grail”: The continued quest to automate
  fact-checking.
\newblock In \emph{Computation+ Journalism Symposium,(September)}.

\bibitem[{AllenNLP(2020)}]{allennlpsrl}
AllenNLP. 2020.
\newblock Allennlp semantic role labeling.
\newblock https://demo.allennlp.org/semantic-role-labeling.
\newblock [Online; accessed 2023-01-02].

\bibitem[{Aly et~al.(2021)Aly, Guo, Schlichtkrull, Thorne, Vlachos,
  Christodoulopoulos, Cocarascu, and
  Mittal}]{https://doi.org/10.48550/arxiv.2106.05707}
Rami Aly, Zhijiang Guo, Michael Schlichtkrull, James Thorne, Andreas Vlachos,
  Christos Christodoulopoulos, Oana Cocarascu, and Arpit Mittal. 2021.
\newblock \href {https://doi.org/10.48550/ARXIV.2106.05707} {Feverous: Fact
  extraction and verification over unstructured and structured information}.

\bibitem[{Atanasova et~al.(2019)Atanasova, Nakov, M{\`a}rquez,
  Barr{\'o}n-Cede{\~n}o, Karadzhov, Mihaylova, Mohtarami, and
  Glass}]{atanasova2019automatic}
Pepa Atanasova, Preslav Nakov, Llu{\'\i}s M{\`a}rquez, Alberto
  Barr{\'o}n-Cede{\~n}o, Georgi Karadzhov, Tsvetomila Mihaylova, Mitra
  Mohtarami, and James Glass. 2019.
\newblock Automatic fact-checking using context and discourse information.
\newblock \emph{Journal of Data and Information Quality (JDIQ)}, 11(3):1--27.

\bibitem[{Bowman et~al.(2015)Bowman, Angeli, Potts, and
  Manning}]{bowman2015large}
Samuel~R Bowman, Gabor Angeli, Christopher Potts, and Christopher~D Manning.
  2015.
\newblock A large annotated corpus for learning natural language inference.
\newblock \emph{arXiv preprint arXiv:1508.05326}.

\bibitem[{Brown et~al.(2020)Brown, Mann, Ryder, Subbiah, Kaplan, Dhariwal,
  Neelakantan, Shyam, Sastry, Askell et~al.}]{brown2020language}
Tom Brown, Benjamin Mann, Nick Ryder, Melanie Subbiah, Jared~D Kaplan, Prafulla
  Dhariwal, Arvind Neelakantan, Pranav Shyam, Girish Sastry, Amanda Askell,
  et~al. 2020.
\newblock Language models are few-shot learners.
\newblock \emph{Advances in neural information processing systems},
  33:1877--1901.

\bibitem[{Devlin et~al.(2018)Devlin, Chang, Lee, and
  Toutanova}]{devlin2018bert}
Jacob Devlin, Ming-Wei Chang, Kenton Lee, and Kristina Toutanova. 2018.
\newblock Bert: Pre-training of deep bidirectional transformers for language
  understanding.
\newblock \emph{arXiv preprint arXiv:1810.04805}.

\bibitem[{Dobbs(2012)}]{dobbs2012rise}
Michael Dobbs. 2012.
\newblock The rise of political fact-checking, how reagan inspired a
  journalistic movement.
\newblock \emph{New America Foundation}, pages 4--5.

\bibitem[{Gao et~al.(2022)Gao, Ma, Lin, and Callan}]{hyde}
Luyu Gao, Xueguang Ma, Jimmy Lin, and Jamie Callan. 2022.
\newblock Precise zero-shot dense retrieval without relevance labels.
\newblock \emph{arXiv preprint arXiv:2212.10496}.

\bibitem[{Garg and Sharma(2020)}]{9337152}
Sonal Garg and Dilip~Kumar Sharma. 2020.
\newblock \href {https://doi.org/10.1109/SMART50582.2020.9337152} {New
  politifact: A dataset for counterfeit news}.
\newblock In \emph{2020 9th International Conference System Modeling and
  Advancement in Research Trends (SMART)}, pages 17--22.

\bibitem[{Guo et~al.(2021)Guo, Schlichtkrull, and
  Vlachos}]{https://doi.org/10.48550/arxiv.2108.11896}
Zhijiang Guo, Michael Schlichtkrull, and Andreas Vlachos. 2021.
\newblock \href {https://doi.org/10.48550/ARXIV.2108.11896} {A survey on
  automated fact-checking}.

\bibitem[{Guo et~al.(2022)Guo, Schlichtkrull, and
  Vlachos}]{10.1162/tacl_a_00454}
Zhijiang Guo, Michael Schlichtkrull, and Andreas Vlachos. 2022.
\newblock \href {https://doi.org/10.1162/tacl_a_00454} {{A Survey on Automated
  Fact-Checking}}.
\newblock \emph{Transactions of the Association for Computational Linguistics},
  10:178--206.

\bibitem[{Gupta and Srikumar(2021)}]{https://doi.org/10.48550/arxiv.2106.09248}
Ashim Gupta and Vivek Srikumar. 2021.
\newblock \href {https://doi.org/10.48550/ARXIV.2106.09248} {X-fact: A new
  benchmark dataset for multilingual fact checking}.

\bibitem[{Hassan et~al.(2015)Hassan, Li, and
  Tremayne}]{10.1145/2806416.2806652}
Naeemul Hassan, Chengkai Li, and Mark Tremayne. 2015.
\newblock \href {https://doi.org/10.1145/2806416.2806652} {Detecting
  check-worthy factual claims in presidential debates}.
\newblock In \emph{Proceedings of the 24th ACM International on Conference on
  Information and Knowledge Management}, CIKM '15, page 1835–1838, New York,
  NY, USA. Association for Computing Machinery.

\bibitem[{Hassan et~al.(2019)Hassan, Hollander, van Lent, and
  Tahoun}]{10.1093/qje/qjz021}
Tarek~A Hassan, Stephan Hollander, Laurence van Lent, and Ahmed Tahoun. 2019.
\newblock \href {https://doi.org/10.1093/qje/qjz021} {{Firm-Level Political
  Risk: Measurement and Effects*}}.
\newblock \emph{The Quarterly Journal of Economics}, 134(4):2135--2202.

\bibitem[{Jiang et~al.(2020)Jiang, Bordia, Zhong, Dognin, Singh, and
  Bansal}]{jiang2020hover}
Yichen Jiang, Shikha Bordia, Zheng Zhong, Charles Dognin, Maneesh Singh, and
  Mohit Bansal. 2020.
\newblock Hover: A dataset for many-hop fact extraction and claim verification.
\newblock \emph{arXiv preprint arXiv:2011.03088}.

\bibitem[{Joshi et~al.(2019)Joshi, Levy, Weld, and
  Zettlemoyer}]{joshi2019coref}
Mandar Joshi, Omer Levy, Daniel~S. Weld, and Luke Zettlemoyer. 2019.
\newblock {BERT} for coreference resolution: Baselines and analysis.
\newblock In \emph{Empirical Methods in Natural Language Processing (EMNLP)}.

\bibitem[{Kumar and Shah(2018)}]{https://doi.org/10.48550/arxiv.1804.08559}
Srijan Kumar and Neil Shah. 2018.
\newblock \href {https://doi.org/10.48550/ARXIV.1804.08559} {False information
  on web and social media: A survey}.

\bibitem[{Kwiatkowski et~al.(2019)Kwiatkowski, Palomaki, Redfield, Collins,
  Parikh, Alberti, Epstein, Polosukhin, Devlin, Lee, Toutanova, Jones, Kelcey,
  Chang, Dai, Uszkoreit, Le, and Petrov}]{kwiatkowski-etal-2019-natural}
Tom Kwiatkowski, Jennimaria Palomaki, Olivia Redfield, Michael Collins, Ankur
  Parikh, Chris Alberti, Danielle Epstein, Illia Polosukhin, Jacob Devlin,
  Kenton Lee, Kristina Toutanova, Llion Jones, Matthew Kelcey, Ming-Wei Chang,
  Andrew~M. Dai, Jakob Uszkoreit, Quoc Le, and Slav Petrov. 2019.
\newblock \href {https://doi.org/10.1162/tacl_a_00276} {Natural questions: A
  benchmark for question answering research}.
\newblock \emph{Transactions of the Association for Computational Linguistics},
  7:452--466.

\bibitem[{Lewis et~al.(2019)Lewis, Liu, Goyal, Ghazvininejad, Mohamed, Levy,
  Stoyanov, and Zettlemoyer}]{DBLP:journals/corr/abs-1910-13461}
Mike Lewis, Yinhan Liu, Naman Goyal, Marjan Ghazvininejad, Abdelrahman Mohamed,
  Omer Levy, Veselin Stoyanov, and Luke Zettlemoyer. 2019.
\newblock \href {http://arxiv.org/abs/1910.13461} {{BART:} denoising
  sequence-to-sequence pre-training for natural language generation,
  translation, and comprehension}.
\newblock \emph{CoRR}, abs/1910.13461.

\bibitem[{Liu et~al.(2019)Liu, Ott, Goyal, Du, Joshi, Chen, Levy, Lewis,
  Zettlemoyer, and Stoyanov}]{liu2019roberta}
Yinhan Liu, Myle Ott, Naman Goyal, Jingfei Du, Mandar Joshi, Danqi Chen, Omer
  Levy, Mike Lewis, Luke Zettlemoyer, and Veselin Stoyanov. 2019.
\newblock Roberta: A robustly optimized bert pretraining approach.
\newblock \emph{arXiv preprint arXiv:1907.11692}.

\bibitem[{Manning et~al.(2014)Manning, Surdeanu, Bauer, Finkel, Bethard, and
  McClosky}]{manning2014stanford}
Christopher~D Manning, Mihai Surdeanu, John Bauer, Jenny~Rose Finkel, Steven
  Bethard, and David McClosky. 2014.
\newblock The stanford corenlp natural language processing toolkit.
\newblock In \emph{Proceedings of 52nd annual meeting of the association for
  computational linguistics: system demonstrations}, pages 55--60.

\bibitem[{Mishra et~al.(2022)Mishra, Suryavardan, Bhaskar, Chopra, Reganti,
  Patwa, Das, Chakraborty, Sheth, Ekbal et~al.}]{mishra2022factify}
Shreyash Mishra, S~Suryavardan, Amrit Bhaskar, Parul Chopra, Aishwarya Reganti,
  Parth Patwa, Amitava Das, Tanmoy Chakraborty, Amit Sheth, Asif Ekbal, et~al.
  2022.
\newblock Factify: A multi-modal fact verification dataset.
\newblock In \emph{Proceedings of the First Workshop on Multimodal
  Fact-Checking and Hate Speech Detection (DE-FACTIFY)}.

\bibitem[{Mott(1942)}]{10.2307/1023893}
Frank~Luther Mott. 1942.
\newblock \href {http://www.jstor.org/stable/1023893} {Trends in newspaper
  content}.
\newblock \emph{The Annals of the American Academy of Political and Social
  Science}, 219:60--65.

\bibitem[{Nakov et~al.(2021)Nakov, Corney, Hasanain, Alam, Elsayed,
  Barr{\'{o}}n{-}Cede{\~{n}}o, Papotti, Shaar, and
  Martino}]{DBLP:journals/corr/abs-2103-07769}
Preslav Nakov, David P.~A. Corney, Maram Hasanain, Firoj Alam, Tamer Elsayed,
  Alberto Barr{\'{o}}n{-}Cede{\~{n}}o, Paolo Papotti, Shaden Shaar, and
  Giovanni Da~San Martino. 2021.
\newblock \href {http://arxiv.org/abs/2103.07769} {Automated fact-checking for
  assisting human fact-checkers}.
\newblock \emph{CoRR}, abs/2103.07769.

\bibitem[{Nicula et~al.(2021)Nicula, Dascalu, Newton, Orcutt, and
  McNamara}]{nicula2021automated}
Bogdan Nicula, Mihai Dascalu, Natalie Newton, Ellen Orcutt, and Danielle~S
  McNamara. 2021.
\newblock Automated paraphrase quality assessment using recurrent neural
  networks and language models.
\newblock In \emph{International Conference on Intelligent Tutoring Systems},
  pages 333--340. Springer.

\bibitem[{Nighojkar and Licato(2021)}]{nighojkar2021improving}
Animesh Nighojkar and John Licato. 2021.
\newblock Improving paraphrase detection with the adversarial paraphrasing
  task.
\newblock \emph{arXiv preprint arXiv:2106.07691}.

\bibitem[{Niu et~al.(2020)Niu, Yavuz, Zhou, Keskar, Wang, and
  Xiong}]{niu2020unsupervised}
Tong Niu, Semih Yavuz, Yingbo Zhou, Nitish~Shirish Keskar, Huan Wang, and
  Caiming Xiong. 2020.
\newblock Unsupervised paraphrasing with pretrained language models.
\newblock \emph{arXiv preprint arXiv:2010.12885}.

\bibitem[{N{\o}rregaard and
  Derczynski(2021)}]{norregaard-derczynski-2021-danfever}
Jeppe N{\o}rregaard and Leon Derczynski. 2021.
\newblock \href {https://aclanthology.org/2021.nodalida-main.47} {{D}an{FEVER}:
  claim verification dataset for {D}anish}.
\newblock In \emph{Proceedings of the 23rd Nordic Conference on Computational
  Linguistics (NoDaLiDa)}, pages 422--428, Reykjavik, Iceland (Online).
  Link{\"o}ping University Electronic Press, Sweden.

\bibitem[{Onoe et~al.(2021)Onoe, Zhang, Choi, and
  Durrett}]{https://doi.org/10.48550/arxiv.2109.01653}
Yasumasa Onoe, Michael J.~Q. Zhang, Eunsol Choi, and Greg Durrett. 2021.
\newblock \href {https://doi.org/10.48550/ARXIV.2109.01653} {Creak: A dataset
  for commonsense reasoning over entity knowledge}.

\bibitem[{Pal et~al.(2022)Pal, Kanoulas, and Rijke}]{pal-etal-2022-parameter}
Vaishali Pal, Evangelos Kanoulas, and Maarten Rijke. 2022.
\newblock \href {https://doi.org/10.18653/v1/2022.dialdoc-1.5}
  {Parameter-efficient abstractive question answering over tables or text}.
\newblock In \emph{Proceedings of the Second DialDoc Workshop on
  Document-grounded Dialogue and Conversational Question Answering}, pages
  41--53, Dublin, Ireland. Association for Computational Linguistics.

\bibitem[{Palmer et~al.(2005)Palmer, Gildea, and
  Kingsbury}]{palmer2005proposition}
Martha Palmer, Daniel Gildea, and Paul Kingsbury. 2005.
\newblock The proposition bank: An annotated corpus of semantic roles.
\newblock \emph{Computational linguistics}, 31(1):71--106.

\bibitem[{Papineni et~al.(2002)Papineni, Roukos, Ward, and
  Zhu}]{papineni2002bleu}
Kishore Papineni, Salim Roukos, Todd Ward, and Wei-Jing Zhu. 2002.
\newblock Bleu: a method for automatic evaluation of machine translation.
\newblock In \emph{Proceedings of the 40th annual meeting of the Association
  for Computational Linguistics}, pages 311--318.

\bibitem[{Park et~al.(2021)Park, Min, Kang, Zettlemoyer, and
  Hajishirzi}]{park2021faviq}
Jungsoo Park, Sewon Min, Jaewoo Kang, Luke Zettlemoyer, and Hannaneh
  Hajishirzi. 2021.
\newblock Faviq: Fact verification from information-seeking questions.
\newblock \emph{arXiv preprint arXiv:2107.02153}.

\bibitem[{Patwa et~al.(2022)Patwa, Mishra, Suryavardan, Bhaskar, Chopra,
  Reganti, Das, Chakraborty, Sheth, Ekbal et~al.}]{patwa2021benchmarking}
Parth Patwa, Shreyash Mishra, S~Suryavardan, Amrit Bhaskar, Parul Chopra,
  Aishwarya Reganti, Amitava Das, Tanmoy Chakraborty, Amit Sheth, Asif Ekbal,
  et~al. 2022.
\newblock Benchmarking multi-modal entailment for fact verification.
\newblock In \emph{Proceedings of De-Factify: Workshop on Multimodal Fact
  Checking and Hate Speech Detection, CEUR}.

\bibitem[{Posetti et~al.(2018)Posetti, Ireton, Wardle, Derakhshan, Matthews,
  Abu-Fadil, Trewinnard, Bell, and Mantzarlis}]{Posetti2018UNESCO}
Julie Posetti, Cherilyn Ireton, Claire Wardle, Hossein Derakhshan, Alice
  Matthews, Magda Abu-Fadil, Tom Trewinnard, Fergus Bell, and Alexios
  Mantzarlis. 2018.
\newblock Unesco.
\newblock https://unesdoc.unesco.org/ark:/48223/pf0000265552.
\newblock [Online; accessed 2023-01-02].

\bibitem[{Qi et~al.(2020)Qi, Yan, Gong, Liu, Duan, Chen, Zhang, and
  Zhou}]{qi-etal-2020-prophetnet}
Weizhen Qi, Yu~Yan, Yeyun Gong, Dayiheng Liu, Nan Duan, Jiusheng Chen, Ruofei
  Zhang, and Ming Zhou. 2020.
\newblock \href {https://doi.org/10.18653/v1/2020.findings-emnlp.217}
  {{P}rophet{N}et: Predicting future n-gram for
  sequence-to-{S}equence{P}re-training}.
\newblock In \emph{Findings of the Association for Computational Linguistics:
  EMNLP 2020}, pages 2401--2410, Online. Association for Computational
  Linguistics.

\bibitem[{Raffel et~al.(2020)Raffel, Shazeer, Roberts, Lee, Narang, Matena,
  Zhou, Li, Liu et~al.}]{raffel2020exploring}
Colin Raffel, Noam Shazeer, Adam Roberts, Katherine Lee, Sharan Narang, Michael
  Matena, Yanqi Zhou, Wei Li, Peter~J Liu, et~al. 2020.
\newblock Exploring the limits of transfer learning with a unified text-to-text
  transformer.
\newblock \emph{J. Mach. Learn. Res.}, 21(140):1--67.

\bibitem[{Schuster et~al.(2021)Schuster, Fisch, and Barzilay}]{schuster2021get}
Tal Schuster, Adam Fisch, and Regina Barzilay. 2021.
\newblock Get your vitamin c! robust fact verification with contrastive
  evidence.
\newblock \emph{arXiv preprint arXiv:2103.08541}.

\bibitem[{Shi and Lin(2019)}]{Shi2019SimpleBM}
Peng Shi and Jimmy Lin. 2019.
\newblock Simple bert models for relation extraction and semantic role
  labeling.
\newblock \emph{ArXiv}, abs/1904.05255.

\bibitem[{Silverman(2020)}]{silverman}
Craig Silverman. 2020.
\newblock \href {https://verificationhandbook.com/} {Verification handbook:
  Homepage}.

\bibitem[{Smarts(2017)}]{smarts_2017}
Media Smarts. 2017.
\newblock \href
  {https://mediasmarts.ca/sites/mediasmarts/files/tip-sheet/tipsheet_false_content.pdf}
  {How to recognize false content online - the new 5 ws}.

\bibitem[{Stofer et~al.(2009)Stofer, Schaffer, and
  Rosenthal}]{stofer2009sports}
Kathryn~T Stofer, James~R Schaffer, and Brian~A Rosenthal. 2009.
\newblock \emph{Sports journalism: An introduction to reporting and writing}.
\newblock Rowman \& Littlefield Publishers.

\bibitem[{Su et~al.(2019)Su, Li, and Wang}]{su2019study}
Jing Su, Xiguang Li, and Lianfeng Wang. 2019.
\newblock The study of a journalism which is almost 99\% fake.
\newblock \emph{Lingue Culture Mediazioni-Languages Cultures Mediation (LCM
  Journal)}, 5(2):115--137.

\bibitem[{Thorne et~al.(2018{\natexlab{a}})Thorne, Vlachos, Christodoulopoulos,
  and Mittal}]{thorne-etal-2018-fever}
James Thorne, Andreas Vlachos, Christos Christodoulopoulos, and Arpit Mittal.
  2018{\natexlab{a}}.
\newblock \href {https://doi.org/10.18653/v1/N18-1074} {{FEVER}: a large-scale
  dataset for fact extraction and {VER}ification}.
\newblock In \emph{Proceedings of the 2018 Conference of the North {A}merican
  Chapter of the Association for Computational Linguistics: Human Language
  Technologies, Volume 1 (Long Papers)}, pages 809--819, New Orleans,
  Louisiana. Association for Computational Linguistics.

\bibitem[{Thorne et~al.(2018{\natexlab{b}})Thorne, Vlachos, Christodoulopoulos,
  and Mittal}]{thorne2018fever}
James Thorne, Andreas Vlachos, Christos Christodoulopoulos, and Arpit Mittal.
  2018{\natexlab{b}}.
\newblock Fever: a large-scale dataset for fact extraction and verification.
\newblock \emph{arXiv preprint arXiv:1803.05355}.

\bibitem[{Trokhymovych and Saez-Trumper(2021)}]{trokhymovych2021wikicheck}
Mykola Trokhymovych and Diego Saez-Trumper. 2021.
\newblock Wikicheck: An end-to-end open source automatic fact-checking api
  based on wikipedia.
\newblock In \emph{Proceedings of the 30th ACM International Conference on
  Information \& Knowledge Management}, pages 4155--4164.

\bibitem[{Wagner and Fischer(1974)}]{wagner1974string}
Robert~A Wagner and Michael~J Fischer. 1974.
\newblock The string-to-string correction problem.
\newblock \emph{Journal of the ACM (JACM)}, 21(1):168--173.

\bibitem[{Wang(2017)}]{wang-2017-liar}
William~Yang Wang. 2017.
\newblock \href {https://doi.org/10.18653/v1/P17-2067} {{``}liar, liar pants on
  fire{''}: A new benchmark dataset for fake news detection}.
\newblock In \emph{Proceedings of the 55th Annual Meeting of the Association
  for Computational Linguistics (Volume 2: Short Papers)}, pages 422--426,
  Vancouver, Canada. Association for Computational Linguistics.

\bibitem[{Wei et~al.(2022)Wei, Wang, Schuurmans, Bosma, Ichter, Xia, Chi, Le,
  and Zhou}]{https://doi.org/10.48550/arxiv.2201.11903}
Jason Wei, Xuezhi Wang, Dale Schuurmans, Maarten Bosma, Brian Ichter, Fei Xia,
  Ed~Chi, Quoc Le, and Denny Zhou. 2022.
\newblock \href {https://doi.org/10.48550/ARXIV.2201.11903} {Chain-of-thought
  prompting elicits reasoning in large language models}.

\bibitem[{Wikipedia(2023)}]{article_2023}
Wikipedia. 2023.
\newblock \href {https://en.wikipedia.org/wiki/Five_Ws} {Five ws}.

\bibitem[{Witteveen and Andrews(2019)}]{witteveen2019paraphrasing}
Sam Witteveen and Martin Andrews. 2019.
\newblock Paraphrasing with large language models.
\newblock \emph{arXiv preprint arXiv:1911.09661}.

\bibitem[{Xun et~al.(2017)Xun, Jia, Gopalakrishnan, and Zhang}]{7580601}
Guangxu Xun, Xiaowei Jia, Vishrawas Gopalakrishnan, and Aidong Zhang. 2017.
\newblock \href {https://doi.org/10.1109/TKDE.2016.2614508} {A survey on
  context learning}.
\newblock \emph{IEEE Transactions on Knowledge and Data Engineering},
  29(1):38--56.

\bibitem[{Yang et~al.(2022{\natexlab{a}})Yang, Vega-Oliveros, Seibt, and
  Rocha}]{yang2022explainable}
Jing Yang, Didier Vega-Oliveros, Ta{\'\i}s Seibt, and Anderson Rocha.
  2022{\natexlab{a}}.
\newblock Explainable fact-checking through question answering.
\newblock In \emph{ICASSP 2022-2022 IEEE International Conference on Acoustics,
  Speech and Signal Processing (ICASSP)}, pages 8952--8956. IEEE.

\bibitem[{Yang et~al.(2022{\natexlab{b}})Yang, Vega-Oliveros, Seibt, and
  Rocha}]{9747214}
Jing Yang, Didier Vega-Oliveros, Taís Seibt, and Anderson Rocha.
  2022{\natexlab{b}}.
\newblock \href {https://doi.org/10.1109/ICASSP43922.2022.9747214} {Explainable
  fact-checking through question answering}.
\newblock In \emph{ICASSP 2022 - 2022 IEEE International Conference on
  Acoustics, Speech and Signal Processing (ICASSP)}, pages 8952--8956.

\bibitem[{Zhang et~al.(2020)Zhang, Zhao, Saleh, and Liu}]{zhang2020pegasus}
Jingqing Zhang, Yao Zhao, Mohammad Saleh, and Peter Liu. 2020.
\newblock Pegasus: Pre-training with extracted gap-sentences for abstractive
  summarization.
\newblock In \emph{International Conference on Machine Learning}, pages
  11328--11339. PMLR.

\bibitem[{Zhao et~al.(2022)Zhao, Arkoudas, Sun, and
  Cardie}]{zhao-etal-2022-compositional}
Wenting Zhao, Konstantine Arkoudas, Weiqi Sun, and Claire Cardie. 2022.
\newblock \href {https://doi.org/10.18653/v1/2022.naacl-main.328}
  {Compositional task-oriented parsing as abstractive question answering}.
\newblock In \emph{Proceedings of the 2022 Conference of the North American
  Chapter of the Association for Computational Linguistics: Human Language
  Technologies}, pages 4418--4427, Seattle, United States. Association for
  Computational Linguistics.

\end{thebibliography}
\bibliographystyle{acl_natbib}
\newpage
\onecolumn
\section{FAQ}

\begin{enumerate}
    \item 5W SRL is understandable, but how is the quality of the 5W QA pair generation using a language model?
    \begin{description}
    \item \textbf{Ans.} - We have evaluated our QA generation against the SoTA model for QA Tasks - T5. Please refer to the section \ref{sec:section7}, table \ref{tab: QAG-QA} for a detailed description of the process and evaluation. Moreover, please see the discussion in the limitation section ~\ref{sec:limitations_qg}.
    \end{description}
    \item How were models shortlisted for Question generation?
    \begin{description}
    \item \textbf{Ans.} - We have shortlisted the current SOTA models on question generation-specific tasks. Due to our resource limitation, we have gone for those models that are open-sourced, are not resource heavy,  and produce great results without fine-tuning them. 
    \end{description}
    \item How were models shortlisted for the question-answering system?
    \begin{description}
    \item \textbf{Ans.} - Selected the current SOTA models that have lower inference time but produce great results on text generation tasks.
    
    \end{description}

    
    \item Why was absolute value $2$ chosen as a filter for minimum edit distance?
    \begin{description}
    \item \textbf{Ans.} - Edit distance is a measure of the similarity between two pieces of text, and a higher value generally indicates more diversity. A higher minimum edit distance between the input and generated text indicates that the generated text is more unique and less likely to be a simple copy or repetition of the input. Therefore, it is commonly held that a minimum edit distance of greater than 2 is a desirable characteristic in natural language generation systems. 
    
    \end{description}

    \item How was the prompt-based paraphrasing done using the \texttt{text-davinci-003} model?
    
    \begin{description}
    \item \textbf{Ans.} - As \texttt{text-davinci-003} is a prompt-based model and so we had to create a prompt that would instruct \texttt{text-davinci-003} to generate five paraphrases for the given input claims. Careful consideration was given to ensure that the prompt would generate output with a specific syntax, as this was necessary for the efficient application of the model to a large number of claims. Through experimentation with multiple different prompts, we came to the conclusion that the following prompt works best:

    \textit{"Generate five different paraphrases of the following text and then place all these five paraphrases in one list of python format. Do not write anything other than just the list "}

    We also developed a post-processing pipeline to ensure that if there is a slight variation in the syntax of paraphrases generated, then we can easily extract those paraphrases from the output of \texttt{text-davinci-003}.

    \end{description}

    \item How was the diversity vs. the number of paraphrases graph plotted?

    \begin{description}
    \item \textbf{Ans.} - After the two layers of filtration, i.e., filtering it by coverage and correctness, the obtained paraphrases are then used to calculate the diversity score as described in section \ref{sec:section3}. Let $d_i$ represent the diversity score of the $i^{th}$ paraphrase generated. So in order to get the general diversity score for the $i^{th}$ paraphrase, we computed the average $d_i$ score of all $i^{th}$ paraphrases generated.
    \end{description}

    
\end{enumerate}

\newpage
\onecolumn
\appendix
\setcounter{section}{0}

\section*{Appendix}\label{sec:appendix}
This section provides additional examples to assist in the understanding and interpretation of the research work presented.
\begin{figure*}[!ht]
\begin{minipage}[b]{0.7\linewidth}
\centering
\resizebox{0.98\textwidth}{!}{%
\begin{tabular}{@{}lllll@{}}
\toprule
\multicolumn{5}{c}{\cellcolor[HTML]{68CBD0}{
\begin{huge}
\textbf{5W QA based Explainability}
\end{huge}
}}
\\
\\ \midrule
\begin{LARGE}\textbf{Who claims}\end{LARGE}     & \begin{LARGE}\textbf{What claims}\end{LARGE}    & \begin{LARGE}\textbf{When claims}\end{LARGE}    & \begin{LARGE}\textbf{Where claims}\end{LARGE}   & \begin{LARGE}\textbf{Why claims}\end{LARGE}    
\\
\hline
\begin{tabular}[c]{@{}l@{}}
\parbox{5cm}{
\begin{Large}
    \centering
    No claim!
\end{Large}
}
\end{tabular} &
  \begin{tabular}[c]{@{}l@{}}
  \parbox{5cm}{
\begin{Large}
  \begin{itemize}
  \item \textbf{Q1}: \textbf{\textit{What is the number of confirmed cases were there in Virginia as of march 18, 2020?}}\\
  \underline{Ans:} More than 77 confirmed cases.\\
  \end{itemize}
\end{Large}
  }
  \end{tabular} &
  \begin{tabular}[c]{@{}l@{}}
  \parbox{5cm}{
\begin{Large}
  \begin{itemize}
  \item \textbf{Q1}: \textbf{\textit{When 77 confirmed cases were reported in the state of Virginia?}}\\ \underline{Ans}: As of March 18 , 2020. \\
  \end{itemize}
\end{Large}
  }
   \end{tabular} &
  \begin{tabular}[c]{@{}l@{}}
  \parbox{5cm}{
\begin{Large}
  \begin{itemize}
  \item \textbf{Q1}: \textbf{\textit{Where were more than 77 confirmed cases reported in 2020?}}\\ \underline{Ans}: In the state of Virginia. \\
  \end{itemize}
\end{Large}

  }
  \end{tabular} &
  \begin{tabular}[c]{@{}l@{}}
  \parbox{5cm}{
\begin{Large}
    \centering
    No claim!
\end{Large}
  }
  \end{tabular}\\
  \hline
\begin{LARGE}\textbf{\textcolor{green}{verified valid}}\end{LARGE}
& 
\includegraphics[width=0.06\textwidth]{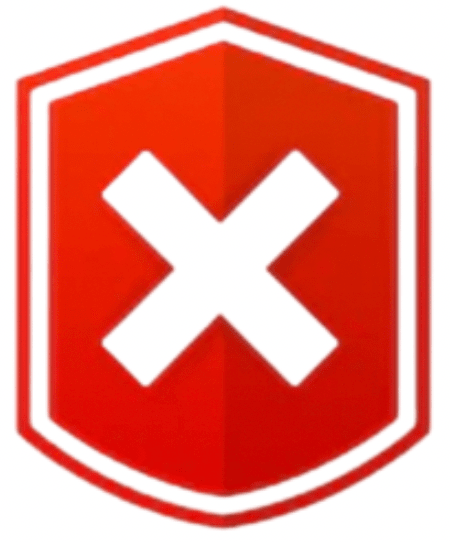} 
\begin{LARGE}\textbf{\textcolor{red}{verified false}}\end{LARGE}
& 
\includegraphics[width=0.06\textwidth]{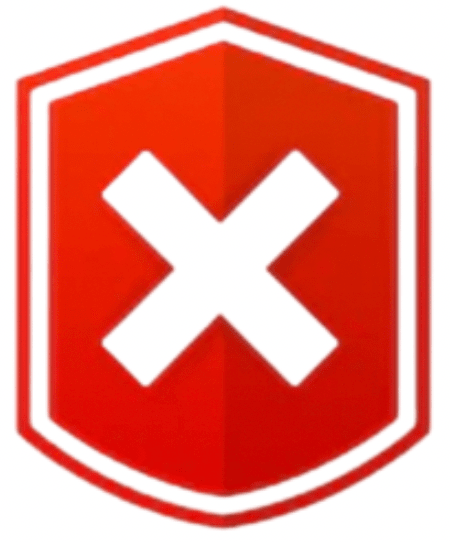} 
\begin{LARGE}\textbf{\textcolor{red}{verified false}}\end{LARGE}
& 
\includegraphics[width=0.04\textwidth]{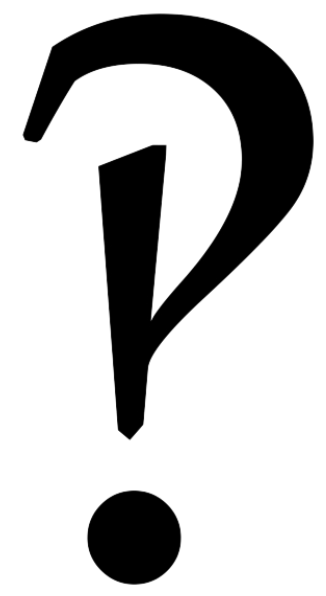} 
\begin{LARGE}\textbf{not verifiable}\end{LARGE}
& 
\includegraphics[width=0.04\textwidth]{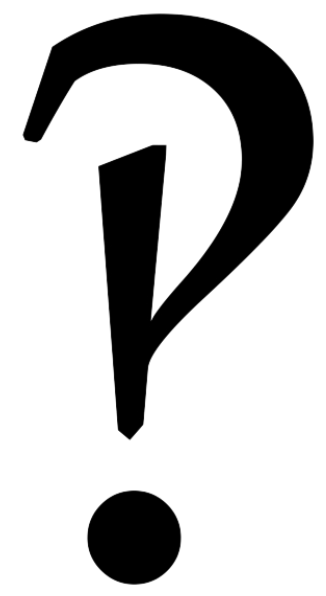} 
\begin{LARGE}\textbf{not verifiable}\end{LARGE}
\\
\hline
\multicolumn{5}{c}{\cellcolor[HTML]{C0C0C0}{
\begin{huge}
\textbf{Evidence}
\end{huge}
}
}  
\\
\hline
\begin{tabular}[c]{@{}l@{}}
\parbox{5cm}{
\begin{Large}
\begin{itemize}
  \item no mention of `who' in any related documents.
\end{itemize}
\end{Large}
}
\end{tabular} &
  \begin{tabular}[c]{@{}l@{}}
  \parbox{5cm}{
\begin{Large}
  \begin{itemize}
  \item The Washington region’s total number of novel coronavirus cases grew to 203 on Wednesday. Maryland added 23 cases Wednesday, bringing the state’s total to 86. \textbf{Virginia reported 10 more cases, for a total of 77}, including the Washington region’s only two deaths.
  \end{itemize}

\end{Large}
  }
  \end{tabular} &
  \begin{tabular}[c]{@{}l@{}}
  \parbox{5cm}{
\begin{Large}
  \begin{itemize}
  \item Virginia has 77 cases of coronavirus as of Wednesday morning,dated \textbf{March 18, 2020}, up an additional 10 cases from the previous day. 
  \end{itemize}
\end{Large}
  }
  \end{tabular} &
  \begin{tabular}[c]{@{}l@{}}
  \parbox{5cm}{
\begin{Large}
  \begin{itemize}
  \item The Washington region’s total number of novel coronavirus cases grew to 203 on Wednesday. Maryland added 23 cases Wednesday, bringing the state’s total to 86. \textbf{Virginia reported 10 more cases, for a total of 77}, including the Washington region’s only two deaths.
  \end{itemize}
\end{Large}
  }
  \end{tabular} &
  \begin{tabular}[c]{@{}l@{}}
  \parbox{5cm}{
\begin{Large}
  \begin{itemize}
  \item no mention of `why' in any related documents. \\
\end{itemize}
\end{Large}
  }\end{tabular} \\
\bottomrule
\end{tabular}%
}
\caption{Claim: As of March 18 , 2020 , there were more than 77 confirmed cases reported in the state of Virginia.}
\label{tab:5WQA_example1}
\end{minipage} \hfill
\begin{minipage}{0.28\linewidth}
\vspace{-8cm}
\centering
\resizebox{\columnwidth}{!}{%
{\tiny
\fbox{%
    \parbox{\columnwidth}{%
    \textcolor{blue}{As of March 18 , 2020 , there were more than 77 confirmed cases reported in the state of Virginia.}
    \\
    \textbf{Prphr 1:} According to records updated on the 18th of March 2020, the state of Virginia has more than 77 COVID-19 cases.
     \\
    \textbf{Prphr 2:} Based on the data of March 18th, 2020, there are over 77 reported cases of coronavirus in Virginia.
     \\
    \textbf{Prphr 3:} By March 18 2020, Virginia has a reported number of more than 77 certified cases of the coronavirus.
     \\
    \textbf{Prphr 4:} As of the 18th of March 2020, there was evidence of 77 positive coronavirus cases in Virginia.   
     \\
    \textbf{Prphr 5:} As of March 18th 2020, 77 documented incidences of coronavirus had been raised in Virginia.  

    }
    }%
    }
}
\vspace{-3mm}
\caption{Claims paraphrased using \texttt{text-davinci-003}}
\label{fig:chatgpt-image1}
\vspace{-3mm}
\end{minipage}
\end{figure*}

\begin{figure*}[!ht]
\begin{minipage}[b]{0.7\linewidth}
\centering
\resizebox{0.98\textwidth}{!}{%
\begin{tabular}{@{}lllll@{}}
\\
\\ \midrule
\begin{LARGE}\textbf{Who claims}\end{LARGE}     & \begin{LARGE}\textbf{What claims}\end{LARGE}    & \begin{LARGE}\textbf{When claims}\end{LARGE}    & \begin{LARGE}\textbf{Where claims}\end{LARGE}   & \begin{LARGE}\textbf{Why claims}\end{LARGE}    
\\
\hline
\begin{tabular}[c]{@{}l@{}}
\parbox{5cm}{
\begin{Large}
  \begin{itemize}
  \item \textbf{Q1}: \textbf{\textit{Who had at least one touchdown pass in each of the first 37 games of the 2014 season?}}\\
  \underline{Ans:} Manning. \\
  \end{itemize}
\end{Large}
}
\end{tabular} &
  \begin{tabular}[c]{@{}l@{}}
  \parbox{5cm}{
\begin{Large}
  \begin{itemize}
  \item \textbf{Q1}: \textbf{\textit{What is the number of touchdown passes did manning have by week 1 of the 2014 season?}}\\
  \underline{Ans:} At least 1 touchdown pass.\\
  \end{itemize}
\end{Large}
  }
  \end{tabular} &
  \begin{tabular}[c]{@{}l@{}}
  \parbox{5cm}{
\begin{Large}
  \begin{itemize}
  \item \textbf{Q1}: \textbf{\textit{When did manning have at least one touchdown pass in all 37 games he played for the broncos?}}\\ \underline{Ans}: By Week 1 of the 2014 season. \\
  \end{itemize}
\end{Large}
  }
   \end{tabular} &
  \begin{tabular}[c]{@{}l@{}}
  \parbox{5cm}{
\begin{Large}
  \begin{itemize}
  \item \textbf{Q1}: \textbf{\textit{Where manning had at least one touchdown pass?}}\\ \underline{Ans}: In the 37 games he has played for the Broncos. \\
  \end{itemize}
\end{Large}

  }
  \end{tabular} &
  \begin{tabular}[c]{@{}l@{}}
  \parbox{5cm}{
\begin{Large}
    \centering
    No claim!
\end{Large}
  }
  \end{tabular}\\
  \hline 
\begin{LARGE}\textbf{\textcolor{green}{verified valid}}\end{LARGE}
& 
\includegraphics[width=0.06\textwidth]{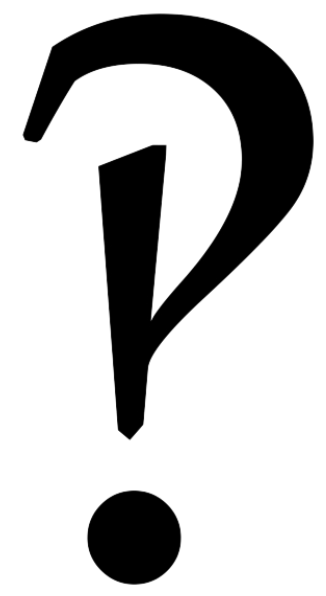} 
\begin{LARGE}\textbf{not verifiable}\end{LARGE}
& 
\begin{LARGE}\textbf{\textcolor{green}{verified valid}}\end{LARGE}
&  
\begin{LARGE}\textbf{\textcolor{green}{verified valid}}\end{LARGE}
& 
\includegraphics[width=0.04\textwidth]{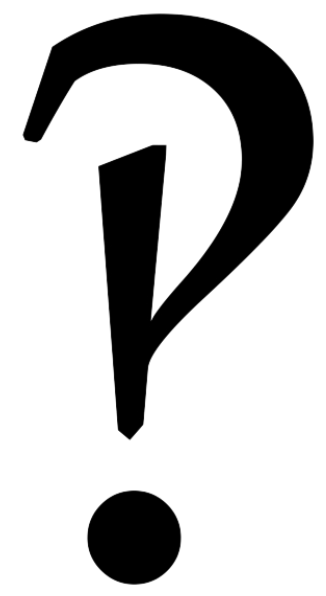} 
\begin{LARGE}\textbf{not verifiable}\end{LARGE}
\\
\hline
\multicolumn{5}{c}{\cellcolor[HTML]{C0C0C0}{
\begin{huge}
\textbf{Evidence}
\end{huge}
}
}  
\\
\hline
\begin{tabular}[c]{@{}l@{}}
\parbox{5cm}{
\begin{Large}
\begin{itemize}
  \item But since arriving in Denver, where he signed a five-year contract that runs through 2016, \textbf{Manning} has somehow been a better version of himself as he adjusted to his new body. He threw 37 touchdowns his first season with the Broncos.
\end{itemize}
\end{Large}
}
\end{tabular} &
  \begin{tabular}[c]{@{}l@{}}
  \parbox{5cm}{
\begin{Large}
  \begin{itemize}
  \item But since arriving in Denver, where he signed a five-year contract that runs through 2016, Manning has somehow been a better version of himself as he adjusted to his new body. He threw 37 touchdowns his first season with the Broncos, while spending more time in the training room and with his doctors than in the weight room as he worked to regain strength in his right triceps and waited for his nerve damage to improve.

  \end{itemize}

\end{Large}
  }
  \end{tabular} &
  \begin{tabular}[c]{@{}l@{}}
  \parbox{5cm}{
\begin{Large}
  \begin{itemize}
  \item \textbf{The Broncos entered the 2014 season} as the defending AFC champions, hoping to compete for another Super Bowl run, following a 43–8 loss to the Seattle Seahawks in Super Bowl XLVIII. 
\item
  
Manning threw a total of 40 touchdown passes, but only four came in the last four games of the regular season and the playoffs. 

  \end{itemize}
\end{Large}
  }
  \end{tabular} &
  \begin{tabular}[c]{@{}l@{}}
  \parbox{5cm}{
\begin{Large}
  \begin{itemize}
  \item He threw 37 touchdowns his first \textbf{season with the Broncos}, while spending more time in the training room and with his doctors than in the weight room as he worked to regain strength in his right triceps and waited for his nerve damage to improve.

  \end{itemize}
\end{Large}
  }
  \end{tabular} &
  \begin{tabular}[c]{@{}l@{}}
  \parbox{5cm}{
\begin{Large}
  \begin{itemize}
  \item no mention of `why' in any related documents. \\
\end{itemize}
\end{Large}
  }\end{tabular} \\
\bottomrule
\end{tabular}%
}
\caption{Claim: By Week 1 of the 2014 season , Manning had at least 1 touchdown pass in the 37 games he has played for the Broncos.}
\label{tab:5WQA_example2}
\end{minipage} \hfill
\begin{minipage}{0.28\linewidth}
\vspace{-8cm}
\centering
\resizebox{\columnwidth}{!}{%
{\tiny
\fbox{%
    \parbox{\columnwidth}{%
    \textcolor{blue}{By Week 1 of the 2014 season, Manning had at least 1 touchdown pass in the 37 games he has played for the Broncos.}
    \\
    \textbf{Prphr 1:} By the kickoff of the 2014 season, Manning had achieved a touchdown pass in 37 of the contests he had featured in for the Broncos.
     \\
    \textbf{Prphr 2:} At the onset of 2014 season Manning had at least one touchdown pass tallied in the 37 competitions participating by the Broncos.
     \\
    \textbf{Prphr 3:} By week 1 of the 2014 season, Manning had tossed over one touchdown pass in the thirty seven contests he participated in for the Broncos.
     \\
    \textbf{Prphr 4:} By the first week of the 2014 season, Manning had a minimum of one touchdown pass in all 37 matches he had played for the Broncos.   
     \\
    \textbf{Prphr 5:} When Week 1 of the 2014 season came around, Manning had attained 1 touchdown pass at least throughout the 37 games he had played for the Broncos.    
    }
    }%
    }
}
\vspace{-3mm}
\caption{Claims paraphrased using \texttt{text-davinci-003}}
\label{fig:chatgpt-image2}
\vspace{-3mm}
\end{minipage}
\end{figure*}

\begin{figure*}[!ht]
\begin{minipage}[b]{0.7\linewidth}
\centering
\resizebox{0.98\textwidth}{!}{%
\begin{tabular}{@{}lllll@{}}
\\
\\ \midrule
\begin{LARGE}\textbf{Who claims}\end{LARGE}     & \begin{LARGE}\textbf{What claims}\end{LARGE}    & \begin{LARGE}\textbf{When claims}\end{LARGE}    & \begin{LARGE}\textbf{Where claims}\end{LARGE}   & \begin{LARGE}\textbf{Why claims}\end{LARGE}    
\\
\hline
\begin{tabular}[c]{@{}l@{}}
\parbox{5cm}{
\begin{Large}
    \centering
    No claim!
\end{Large}
}
\end{tabular} &
  \begin{tabular}[c]{@{}l@{}}
  \parbox{5cm}{
\begin{Large}
  \begin{itemize}
  \item \textbf{Q1}: \textbf{\textit{What is controversial about city morgues music videos?}}\\
  \underline{Ans:} Heavy use of drugs , violence , firearms , and nudity.\\
  \end{itemize}
\end{Large}
  }
  \end{tabular} &
  \begin{tabular}[c]{@{}l@{}}
  \parbox{5cm}{
\begin{Large}
    \centering
    No claim!
\end{Large}
  }
   \end{tabular} &
  \begin{tabular}[c]{@{}l@{}}
  \parbox{5cm}{
\begin{Large}
    \centering
    No claim!
\end{Large}
  }
  \end{tabular} &
  \begin{tabular}[c]{@{}l@{}}
  \parbox{5cm}{
\begin{Large}
  \begin{itemize}
  \item \textbf{Q1}: \textbf{\textit{Why are city morgue's music videos controversial?}}\\ \underline{Ans}: As they show heavy use of drugs , violence , firearms , and nudity. \\
  \end{itemize}
\end{Large}
  }
  \end{tabular}\\
  \hline
\includegraphics[width=0.04\textwidth]{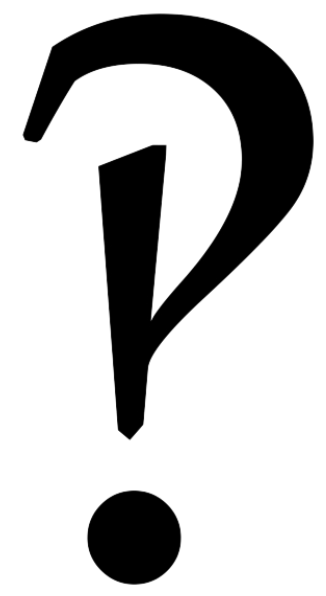} 
\begin{LARGE}\textbf{not verifiable}\end{LARGE}
& 
\begin{LARGE}\textbf{\textcolor{green}{verified valid}}\end{LARGE}
& 
\includegraphics[width=0.06\textwidth]{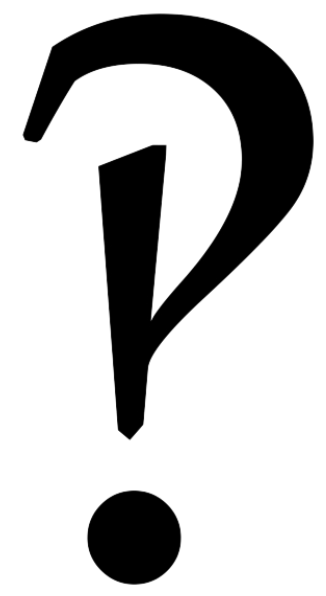} 
\begin{LARGE}\textbf{not verifiable}\end{LARGE}
& 
\includegraphics[width=0.04\textwidth]{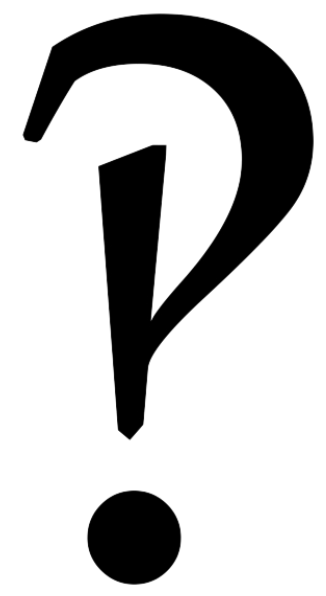} 
\begin{LARGE}\textbf{not verifiable}\end{LARGE}
& 
\begin{LARGE}\textbf{\textcolor{green}{verified valid}}\end{LARGE}
\\
\hline
\multicolumn{5}{c}{\cellcolor[HTML]{C0C0C0}{
\begin{huge}
\textbf{Evidence}
\end{huge}
}
}  
\\
\hline
\begin{tabular}[c]{@{}l@{}}
\parbox{5cm}{
\begin{Large}
\begin{itemize}
\item no mention of `who' in any related documents. 
\end{itemize}
\end{Large}
}
\end{tabular} &
  \begin{tabular}[c]{@{}l@{}}
  \parbox{5cm}{
\begin{Large}
  \begin{itemize}
  \item City Morgue is an American hip hop group from New York, best known for their controversial music videos depicting the \textbf{heavy use of narcotics, violence, weaponry (mainly firearms), and nudity.}

  \end{itemize}

\end{Large}
  }
  \end{tabular} &
  \begin{tabular}[c]{@{}l@{}}
  \parbox{5cm}{
\begin{Large}
  \begin{itemize}
\item no mention of `when' in any related documents. 

  \end{itemize}
\end{Large}
  }
  \end{tabular} &
  \begin{tabular}[c]{@{}l@{}}
  \parbox{5cm}{
\begin{Large}
  \begin{itemize}
\item no mention of `where' in any related documents.

  \end{itemize}
\end{Large}
  }
  \end{tabular} &
  \begin{tabular}[c]{@{}l@{}}
  \parbox{5cm}{
\begin{Large}
  \begin{itemize}
  \item City Morgue is an American hip hop group from New York, best known for their controversial music videos depicting the \textbf{heavy use of narcotics, violence, weaponry (mainly firearms), and nudity.} \\
\end{itemize}
\end{Large}
  }\end{tabular} \\
\bottomrule
\end{tabular}%
}
\caption{Claim: City Morgue s music videos are controversial as they show heavy use of drugs , violence , firearms , and nudity.}
\label{tab:5WQA_example2_1}
\end{minipage} \hfill
\begin{minipage}{0.28\linewidth}
\vspace{-8cm}
\centering
\resizebox{\columnwidth}{!}{%
{\tiny
\fbox{%
    \parbox{\columnwidth}{%
    \textcolor{blue}{City Morgue s music videos are controversial as they show heavy use of drugs , violence , firearms , and nudity.}
    \\
    \textbf{Prphr 1:} City Morgue's song visuals are controversial as they present hefty utilization of drugs, savagery, firearms, and nakedness.
     \\
    \textbf{Prphr 2:} City Morgue's music videos are seen as disputable due to their extensive portrayal of drug usage, brutality, weaponry, and nudity.
     \\
    \textbf{Prphr 3:} City Morgue's music video content has caused debate for its graphically demonstrating of narcotics, brutality, firearms, and stark nudity.
     \\
    \textbf{Prphr 4:} City Morgue has become notorious for the contentiousness of their music videos due to its frank exhibition of drugs, violence, guns, and nudity.  
     \\
    \textbf{Prphr 5:} City Morgue's music videos have been deemed controversial due to the inclusion of drugs, violence, guns, and nudity.  
    }
    }%
    }
}
\vspace{-3mm}
\caption{Claims paraphrased using \texttt{text-davinci-003}}
\label{fig:chatgpt-image3}
\vspace{-3mm}
\end{minipage}
\end{figure*}

\begin{figure*}[!ht]
\begin{minipage}[b]{0.7\linewidth}
\centering
\resizebox{0.98\textwidth}{!}{%
\begin{tabular}{@{}lllll@{}}
\\
\\ \midrule
\begin{LARGE}\textbf{Who claims}\end{LARGE}     & \begin{LARGE}\textbf{What claims}\end{LARGE}    & \begin{LARGE}\textbf{When claims}\end{LARGE}    & \begin{LARGE}\textbf{Where claims}\end{LARGE}   & \begin{LARGE}\textbf{Why claims}\end{LARGE}    
\\
\hline
\begin{tabular}[c]{@{}l@{}}
\parbox{5cm}{
\begin{Large}
    \centering
    No claim!
\end{Large}
}
\end{tabular} &
  \begin{tabular}[c]{@{}l@{}}
  \parbox{5cm}{
\begin{Large}
  \begin{itemize}
  \item \textbf{Q1}: \textbf{\textit{What movie was nominated for best animated feature and best original score?}}\\
  \underline{Ans}: How to Train Your Dragon.\\
  \end{itemize}
\end{Large}
  }
  \end{tabular} &
  \begin{tabular}[c]{@{}l@{}}
  \parbox{5cm}{
\begin{Large}
    \centering
    No claim!
\end{Large}
  }
   \end{tabular} &
  \begin{tabular}[c]{@{}l@{}}
  \parbox{5cm}{
\begin{Large}
  \begin{itemize}
  \item \textbf{Q1}: \textbf{\textit{Where was how to train your dragon nominated for an academy award?}}\\ \underline{Ans}: At the 83rd Academy Awards. \\
  \end{itemize}
\end{Large}
  }
  \end{tabular} &
  \begin{tabular}[c]{@{}l@{}}
  \parbox{5cm}{
\begin{Large}
    \centering
    No claim!
\end{Large}
  }
  \end{tabular}\\
  \hline
\includegraphics[width=0.04\textwidth]{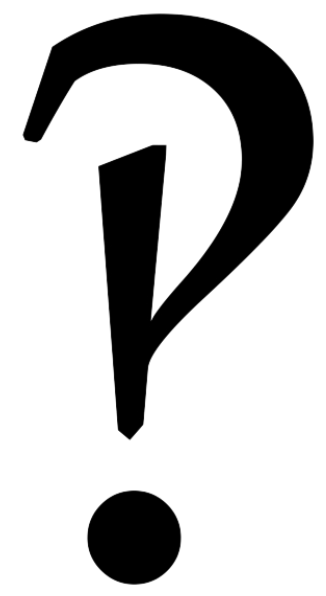} 
\begin{LARGE}\textbf{{not verifiable}}\end{LARGE}
& 
\begin{LARGE}\textbf{\textcolor{green}{verified valid}}\end{LARGE}
& 
\includegraphics[width=0.06\textwidth]{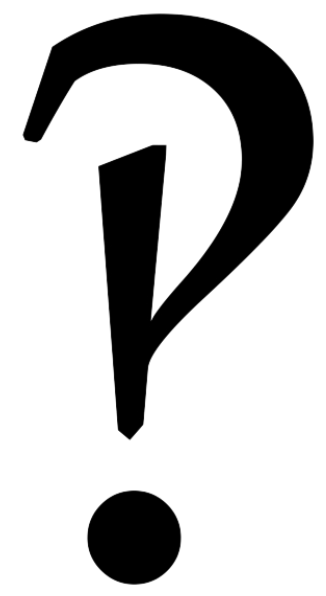} 
\begin{LARGE}\textbf{{not verifiable}}\end{LARGE}
& 
\begin{LARGE}\textbf{\textcolor{green}{verified valid}}\end{LARGE}
& 
\includegraphics[width=0.04\textwidth]{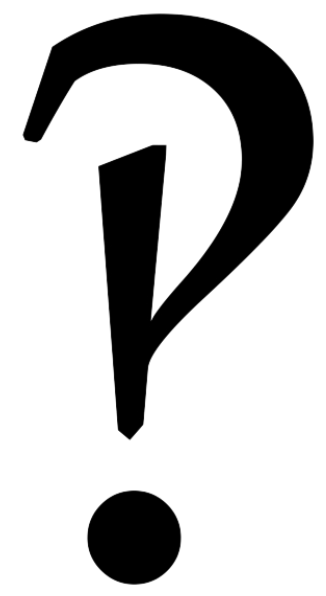} 
\begin{LARGE}\textbf{not verifiable}\end{LARGE}
\\
\hline
\multicolumn{5}{c}{\cellcolor[HTML]{C0C0C0}{
\begin{huge}
\textbf{Evidence}
\end{huge}
}
}  
\\
\hline
\begin{tabular}[c]{@{}l@{}}
\parbox{5cm}{
\begin{Large}
\begin{itemize}
\item no mention of `who' in any related documents 
\end{itemize}
\end{Large}
}
\end{tabular} &
  \begin{tabular}[c]{@{}l@{}}
  \parbox{5cm}{
\begin{Large}
  \begin{itemize}
  \item \textbf{How to Train Your Dragon} premiered at the Gibson Amphitheater on March 21, 2010, and was released in the United States five days later on March 26. The film was a commercial success, earning nearly \$500 million worldwide. It was widely acclaimed, being praised for its animation, voice acting, writing, musical score, and 3D sequences. It was nominated for the Academy Award for Best Animated Feature and Best Original Score at the 83rd Academy Awards, but lost to Toy Story 3 and The Social Network, respectively.

  \end{itemize}

\end{Large}
  }
  \end{tabular} &
  \begin{tabular}[c]{@{}l@{}}
  \parbox{5cm}{
\begin{Large}
  \begin{itemize}
\item no mention of `when' in any related documents. 

  \end{itemize}
\end{Large}
  }
  \end{tabular} &
  \begin{tabular}[c]{@{}l@{}}
  \parbox{5cm}{
\begin{Large}
  \begin{itemize}
\item How to Train Your Dragon premiered at the Gibson Amphitheater on March 21, 2010, and was released in the United States five days later on March 26. The film was a commercial success, earning nearly \$500 million worldwide. It was widely acclaimed, being praised for its animation, voice acting, writing, musical score, and 3D sequences. It was nominated for the Academy Award for Best Animated Feature and Best Original Score at \textbf{the 83rd Academy Awards}, but lost to Toy Story 3 and The Social Network, respectively.

  \end{itemize}
\end{Large}
  }
  \end{tabular} &
  \begin{tabular}[c]{@{}l@{}}
  \parbox{5cm}{
\begin{Large}
  \begin{itemize}
  \item no mention of `why' in any related documents. \\
\end{itemize}
\end{Large}
  }\end{tabular} \\
\bottomrule
\end{tabular}%
}
\caption{Claim: How to Train Your Dragon was nominated for the Academy Award for Best Animated Feature and Best Original Score at the 83rd Academy Awards.}
\label{tab:5WQA_example4}
\end{minipage} \hfill
\begin{minipage}{0.28\linewidth}
\vspace{-8cm}
\centering
\resizebox{\columnwidth}{!}{%
{\tiny
\fbox{%
    \parbox{\columnwidth}{%
    \textcolor{blue}{How to Train Your Dragon was nominated for the Academy Award for Best Animated Feature and Best Original Score at the 83rd Academy Awards .}
    \\
    \textbf{Prphr 1:} The Academy Award was made to How to Train Your Dragon for Best Animated Feature and Best Original Score at the 83rd Academy Awards.
     \\
    \textbf{Prphr 2:} How to Train Your Dragon earned a nomination for the Academy Award for Best Animated Feature and Best Original Score at the 83rd Academy Awards.
     \\
    \textbf{Prphr 3:} How to Train Your Dragon got selected for the Academy Award for Best Animated Feature and Best Original Score at the 83rd Academy Awards.
     \\
    \textbf{Prphr 4:} How to Train Your Dragon was put forward for the Academy Award for Best Animated Feature and Best Original Score at the 83rd Academy Awards.   
     \\
    \textbf{Prphr 5:} At the 83rd Academy Awards, the nomination of How to Train Your Dragon was bagged in the Best Animated Feature and Best Original Score categories.
    }
    }%
    }
}
\vspace{-3mm}
\caption{Claims paraphrased using \texttt{text-davinci-003}}
\label{fig:chatgpt-image4}
\vspace{-3mm}
\end{minipage}
\end{figure*}

\begin{figure*}[!ht]
\begin{minipage}[b]{0.7\linewidth}
\centering
\resizebox{0.98\textwidth}{!}{%
\begin{tabular}{@{}lllll@{}}
\\
\\ \midrule
\begin{LARGE}\textbf{Who claims}\end{LARGE}     & \begin{LARGE}\textbf{What claims}\end{LARGE}    & \begin{LARGE}\textbf{When claims}\end{LARGE}    & \begin{LARGE}\textbf{Where claims}\end{LARGE}   & \begin{LARGE}\textbf{Why claims}\end{LARGE}    
\\
\hline
\begin{tabular}[c]{@{}l@{}}
\parbox{5cm}{
\begin{Large}
  \begin{itemize}
  \item \textbf{Q1}: \textbf{\textit{Who was sent off in the 42nd minute at manchester city?}}\\
  \underline{Ans:} Medhi Benatia. \\
  \end{itemize}
\end{Large}
}
\end{tabular} &
  \begin{tabular}[c]{@{}l@{}}
  \parbox{5cm}{
\begin{Large}
  \begin{itemize}
    \centering
    \item No claim!
  \end{itemize}
\end{Large}
  }
  \end{tabular} &
  \begin{tabular}[c]{@{}l@{}}
  \parbox{5cm}{
\begin{Large}
  \begin{itemize}
  \item \textbf{Q1}: \textbf{\textit{When was medhi benatia sent off?}}\\ \underline{Ans}: In the 42nd minute. \\
  \end{itemize}
\end{Large}
  }
   \end{tabular} &
  \begin{tabular}[c]{@{}l@{}}
  \parbox{5cm}{
\begin{Large}
  \begin{itemize}
    \centering
    \item No claim!
  \end{itemize}
\end{Large}
  }
  \end{tabular} &
  \begin{tabular}[c]{@{}l@{}}
  \parbox{5cm}{
\begin{Large}
 \begin{itemize}
  \item \textbf{Q1}: \textbf{\textit{Why was medhi benatia sent off?}}\\ \underline{Ans}: For an infraction on Fernandinho. \\
  \end{itemize}
\end{Large}
  }
  \end{tabular}\\
  \hline
\begin{LARGE}\textbf{\textcolor{green}{verified valid}}\end{LARGE}
& 
\includegraphics[width=0.06\textwidth]{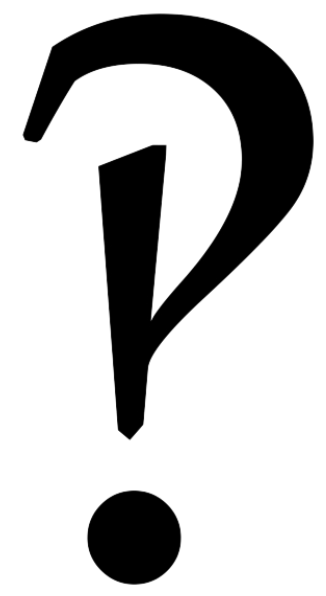} 
\begin{LARGE}\textbf{not verifiable}\end{LARGE}
& 
\includegraphics[width=0.06\textwidth]{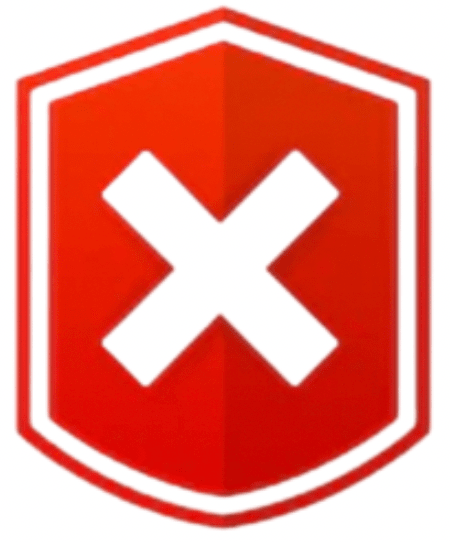} 
\begin{LARGE}\textbf{\textcolor{red}{verified false}}\end{LARGE}
& 
\includegraphics[width=0.04\textwidth]{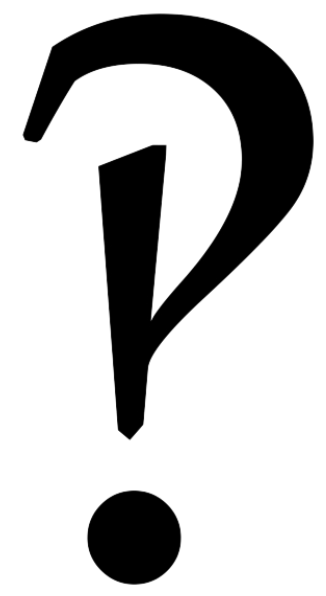} 
\begin{LARGE}\textbf{not verifiable}\end{LARGE}
&  
\includegraphics[width=0.06\textwidth]{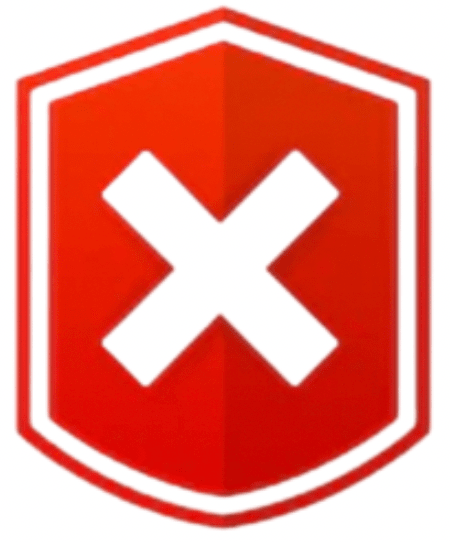} 
\begin{LARGE}\textbf{\textcolor{red}{verified false}}\end{LARGE}
\\
\hline
\multicolumn{5}{c}{\cellcolor[HTML]{C0C0C0}{
\begin{huge}
\textbf{Evidence}
\end{huge}
}
}  
\\
\hline
\begin{tabular}[c]{@{}l@{}}
\parbox{5cm}{
\begin{Large}
\begin{itemize}
  \item On 17 September 2014, \textbf{Benatia} made his official debut for Bayern in a 1–0 home win against Manchester City, for the opening match of the 2014–15 UEFA Champions League season, where he played for 85 minutes, completing 93\% of his passes. In the return match at Manchester City, he was sent off in the 20th minute for denying Sergio Agüero a clear goalscoring opportunity
\end{itemize}
\end{Large}
}
\end{tabular} &
  \begin{tabular}[c]{@{}l@{}}
  \parbox{5cm}{
\begin{Large}
  \begin{itemize}
  \item no mention of 'what' in any related documents.

  \end{itemize}

\end{Large}
  }
  \end{tabular} &
  \begin{tabular}[c]{@{}l@{}}
  \parbox{5cm}{
\begin{Large}
  \begin{itemize}
  \item In the return match at Manchester City , he was sent off in the \textbf{20th minute} for denying Sergio Agüero a clear goalscoring opportunity ; the subsequent penalty was converted by Agüero and City went on to win 3–2.Benatia scored his first goal for Bayern on 13 December , opening the scoring in a 4–0 win at FC Augsburg with a header ; this result put his club 10 points clear at the top of the Bundesliga table.

  \end{itemize}
\end{Large}
  }
  \end{tabular} &
  \begin{tabular}[c]{@{}l@{}}
  \parbox{5cm}{
\begin{Large}
  \begin{itemize}
  \item \textbf{In the return match at Manchester City} , he was sent off in the 20th minute for denying Sergio Agüero a clear goalscoring opportunity.

  \end{itemize}
\end{Large}
  }
  \end{tabular} &
  \begin{tabular}[c]{@{}l@{}}
  \parbox{5cm}{
\begin{Large}
  \begin{itemize}
  \item In the return match at Manchester City , \textbf{he was sent off in the 20th minute for denying Sergio Agüero a clear goalscoring opportunity} ; the subsequent penalty was converted by Agüero and City went on to win 3–2.Benatia scored his first goal for Bayern on 13 December , opening the scoring in a 4–0 win at FC Augsburg with a header ; this result put his club 10 points clear at the top of the Bundesliga table.\\
\end{itemize}
\end{Large}
  }\end{tabular} \\ \bottomrule

\end{tabular}%
}
\caption{Claim: In the return match at Manchester City , Medhi Benatia was sent off in the 42nd minute for an infraction on Fernandinho.}
\label{tab:5WQA_example5}
\end{minipage} \hfill
\begin{minipage}{0.28\linewidth}
\vspace{-8cm}
\centering
\resizebox{\columnwidth}{!}{%
{\tiny
\fbox{%
    \parbox{\columnwidth}{%
    \textcolor{blue}{In the return match at Manchester City , Medhi Benatia was sent off in the 42nd minute for an infraction on Fernandinho.}
    \\
    \textbf{Prphr 1:} In the rematch conducted at Manchester City, Mehi Benatia was dismissed in the 42nd minute as he committed a foul towards Fernandinho.
     \\
    \textbf{Prphr 2:} Back at Manchester City for the return game, Mehi Benatia was penalized with a red card in the 42nd minute for the infraction on Fernandinho.
     \\
    \textbf{Prphr 3:} In the game held again at Manchester City, Medhi Benatia got his marching orders in the 42nd minute for a foul on Fernandinho.
     \\
    \textbf{Prphr 4:} In the game held again in Manchester City, Medhi Benatia got a red card in the 42nd minute due to an infraction on Fernandinho.   
     \\
    \textbf{Prphr 5:} It was in Manchester City for the rematch when Medhi Benatia was shown the red card in the 42nd minute as a consequence of a grave infraction on Fernandinho.    
    }
    }%
    }
}
\vspace{-3mm}
\caption{Claims paraphrased using \texttt{text-davinci-003}}
\label{fig:chatgpt-image5}
\vspace{-3mm}
\end{minipage}
\end{figure*}

\begin{figure*}[!ht]
\begin{minipage}[b]{0.7\linewidth}
\centering
\resizebox{0.98\textwidth}{!}{%
\begin{tabular}{@{}lllll@{}}
\\
\\ \midrule
\begin{LARGE}\textbf{Who claims}\end{LARGE}     & \begin{LARGE}\textbf{What claims}\end{LARGE}    & \begin{LARGE}\textbf{When claims}\end{LARGE}    & \begin{LARGE}\textbf{Where claims}\end{LARGE}   & \begin{LARGE}\textbf{Why claims}\end{LARGE}    
\\
\hline
\begin{tabular}[c]{@{}l@{}}
\parbox{5cm}{
\begin{Large}
  \begin{itemize}
  \item \textbf{Q1}: \textbf{\textit{Who produced avengers assemble?
}}\\
  \underline{Ans}: The director of action movie Batman : Mask of the Phantasm.\\
  \end{itemize}
\end{Large}
}
\end{tabular} &
  \begin{tabular}[c]{@{}l@{}}
  \parbox{5cm}{
\begin{Large}
  \begin{itemize}
  \item \textbf{Q1}: \textbf{\textit{what was the name of the movie produced by batman : mask of the phantasm director?
}}\\
  \underline{Ans}: Avengers Assemble.\\
  \end{itemize}
\end{Large}
  }
  \end{tabular} &
  \begin{tabular}[c]{@{}l@{}}
  \parbox{5cm}{
\begin{Large}
  \begin{itemize}
  \item \textbf{Q1}: \textbf{\textit{When did avengers assemble premiere?}}\\
  \underline{Ans}: On May 26 , 2013.\\
  \end{itemize}
\end{Large}
  }
   \end{tabular} &
  \begin{tabular}[c]{@{}l@{}}
  \parbox{5cm}{
\begin{Large}
  \begin{itemize}
  \item \textbf{Q1}: \textbf{\textit{Where did avengers assemble premiere?}}\\ \underline{Ans}: On Disney XD. \\
  \end{itemize}
\end{Large}
  }
  \end{tabular} &
  \begin{tabular}[c]{@{}l@{}}
  \parbox{5cm}{
\begin{Large}
    \centering
    No claim!
\end{Large}
  }
  \end{tabular}\\
  \hline
\begin{LARGE}\textbf{\textcolor{green}{verified valid}}\end{LARGE}
& 
\begin{LARGE}\textbf{\textcolor{green}{verified valid}}\end{LARGE}
& 
\begin{LARGE}\textbf{\textcolor{green}{verified valid}}\end{LARGE}
& 
\begin{LARGE}\textbf{\textcolor{green}{verified valid}}\end{LARGE}
& 
\includegraphics[width=0.04\textwidth]{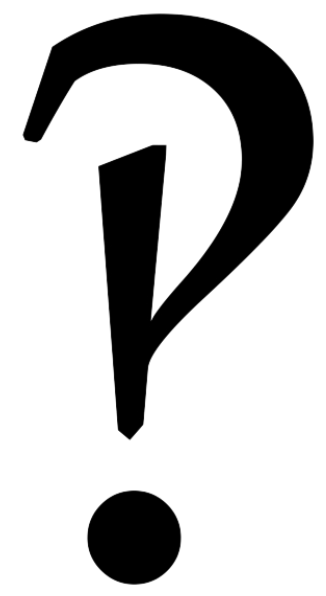} 
\begin{LARGE}\textbf{not verifiable}\end{LARGE}
\\
\hline
\multicolumn{5}{c}{\cellcolor[HTML]{C0C0C0}{
\begin{huge}
\textbf{Evidence}
\end{huge}
}
}  
\\
\hline
\begin{tabular}[c]{@{}l@{}}
\parbox{5cm}{
\begin{Large}
\begin{itemize}
\item \textbf{Eric Radomsky} is one of the producers and directors of Avengers Assemble. He is also the Marvel Animation's Senior Vice President and Creative Director of Animation. He is perhaps best known as co-creator and co-producer of the Emmy award-winning Batman: Mask of the Phantasm. 
\end{itemize}
\end{Large}
}
\end{tabular} &
  \begin{tabular}[c]{@{}l@{}}
  \parbox{5cm}{
\begin{Large}
  \begin{itemize}
  \item Eric Radomsky is one of the producers and directors of \textbf{Avengers Assemble}. He is also the Marvel Animation's Senior Vice President and Creative Director of Animation. He is perhaps best known as co-creator and co-producer of the Emmy award-winning Batman: Mask of the Phantasm.

  \end{itemize}

\end{Large}
  }
  \end{tabular} &
  \begin{tabular}[c]{@{}l@{}}
  \parbox{5cm}{
\begin{Large}
  \begin{itemize}
\item \textbf{M.O.D.O.K. Avengers Assemble} is an animated series, based on the fictional Marvel Comics superhero team the Avengers, which has been designed to capitalize on the success of The Avengers. Avengers Assemble premiered on May 26, 2013, on Disney XD.

  \end{itemize}
\end{Large}
  }
  \end{tabular} &
  \begin{tabular}[c]{@{}l@{}}
  \parbox{5cm}{
\begin{Large}
  \begin{itemize}
\item M.O.D.O.K. Avengers Assemble is an animated series, based on the fictional Marvel Comics superhero team the Avengers, which has been designed to capitalize on the success of The Avengers. Avengers Assemble premiered on May 26, 2013, on \textbf{Disney XD}.

  \end{itemize}
\end{Large}
  }
  \end{tabular} &
  \begin{tabular}[c]{@{}l@{}}
  \parbox{5cm}{
\begin{Large}
  \begin{itemize}
  \item no mention of `why' in any related documents. \\
\end{itemize}
\end{Large}
  }\end{tabular} \\
\bottomrule
\end{tabular}%
}
\caption{Claim: The director of action movie Batman: Mask of the Phantasm, produced Avengers Assemble that premiered on Disney XD on May 26, 2013.}
\label{tab:5WQA_example6}
\end{minipage} \hfill
\begin{minipage}{0.28\linewidth}
\vspace{-8cm}
\centering
\resizebox{\columnwidth}{!}{%
{\tiny
\fbox{%
    \parbox{\columnwidth}{%
    \textcolor{blue}{The director of action movie Batman: Mask of the Phantasm, produced Avengers Assemble that premiered on Disney XD on May 26, 2013 .}
    \\
    \textbf{Prphr 1:} The director of Batman: Mask of the Phantasm, which is an action flick, created Avengers Assemble and it made its premiere on Disney XD on May 26th, 2013.
     \\
    \textbf{Prphr 2:} The director of the action-thriller Batman: Mask of the Phantasm authored Avengers Assemble premiering on the Disney XD portal on 26 May 2013.
     \\
    \textbf{Prphr 3:} The director behind the action movie Batman: Mask of the Phantasm gave birth to Avengers Assemble viewed on Disney XD 26th May 2013.
     \\
    \textbf{Prphr 4:} The Batman: Mask of the Phantasm director was also responsible for Avengers Assemble which debuted on Disney XD on 26/05/2013.   
     \\
    \textbf{Prphr 5:} The person who worked as the director for the action movie Batman: Mask of the Phantasm made Avengers Assemble and it was first aired on Disney XD on May 26th 2013. 
    }
    }%
    }
}
\vspace{-3mm}
\caption{Claims paraphrased using \texttt{text-davinci-003}}
\label{fig:chatgpt-image6}
\vspace{-3mm}
\end{minipage}
\end{figure*}

\begin{figure*}[!ht]
\begin{minipage}[b]{0.7\linewidth}
\centering
\resizebox{0.98\textwidth}{!}{%
\begin{tabular}{@{}lllll@{}}
\\
\\ \midrule
\begin{LARGE}\textbf{Who claims}\end{LARGE}     & \begin{LARGE}\textbf{What claims}\end{LARGE}    & \begin{LARGE}\textbf{When claims}\end{LARGE}    & \begin{LARGE}\textbf{Where claims}\end{LARGE}   & \begin{LARGE}\textbf{Why claims}\end{LARGE}    
\\
\hline
\begin{tabular}[c]{@{}l@{}}
\parbox{5cm}{
\begin{Large}
  \begin{itemize}
  \item \textbf{Q1}: \textbf{\textit{Who was benched during houston's game against texas tech?
}}\\
  \underline{Ans}: Allen.\\
  \end{itemize}
\end{Large}
}
\end{tabular} &
  \begin{tabular}[c]{@{}l@{}}
  \parbox{5cm}{
\begin{Large}
    \centering
    No claim!
\end{Large}
  }
  \end{tabular} &
  \begin{tabular}[c]{@{}l@{}}
  \parbox{5cm}{
\begin{Large}
  \begin{itemize}
  \item \textbf{Q1}: \textbf{\textit{when was allen benched?}}\\
  \underline{Ans}: During Houstons game against Texas Tech.\\
  \end{itemize}
\end{Large}
  }
   \end{tabular} &
  \begin{tabular}[c]{@{}l@{}}
  \parbox{5cm}{
\begin{Large}
    \centering
    No claim!
\end{Large}
  }
  \end{tabular} &
  \begin{tabular}[c]{@{}l@{}}
  \parbox{5cm}{
\begin{Large}
    \centering
    No claim!
\end{Large}
  }
  \end{tabular}\\
  \hline
\begin{LARGE}\textbf{\textcolor{green}{verified valid}}\end{LARGE}
& 
\includegraphics[width=0.06\textwidth]{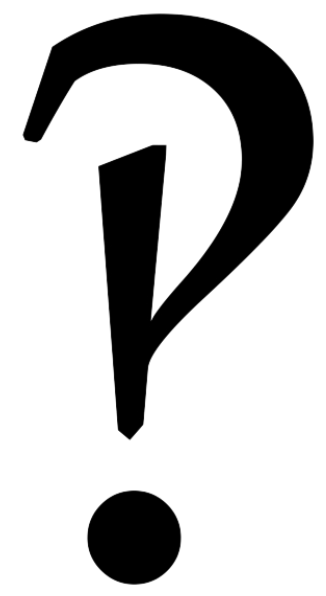} 
\begin{LARGE}\textbf{{not verifiable}}\end{LARGE}
& 
\begin{LARGE}\textbf{\textcolor{green}{verified valid}}\end{LARGE}
& 
\includegraphics[width=0.04\textwidth]{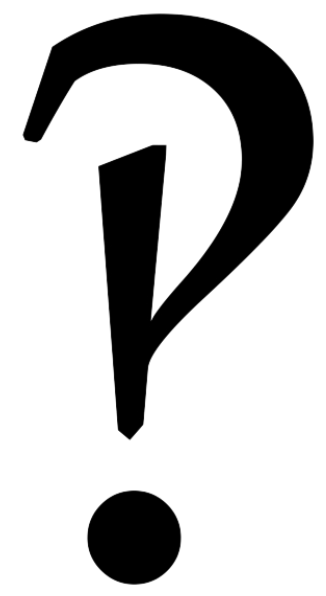} 
\begin{LARGE}\textbf{not verifiable}\end{LARGE}
& 
\includegraphics[width=0.04\textwidth]{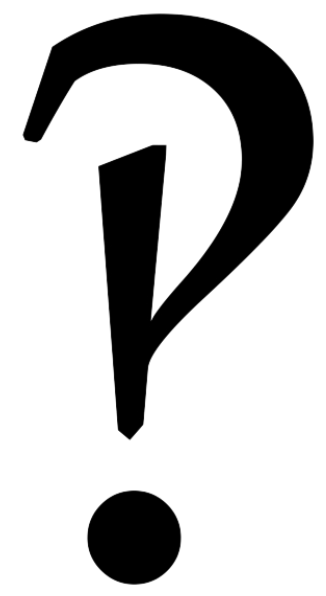} 
\begin{LARGE}\textbf{not verifiable}\end{LARGE}
\\
\hline
\multicolumn{5}{c}{\cellcolor[HTML]{C0C0C0}{
\begin{huge}
\textbf{Evidence}
\end{huge}
}
}  
\\
\hline
\begin{tabular}[c]{@{}l@{}}
\parbox{5cm}{
\begin{Large}
\begin{itemize}
\item \textbf{Kyle Allen} began last season as UH's starting quarterback, but he was benched in a loss to Texas Tech and only play briefly the remainder of the year.
\end{itemize}
\end{Large}
}
\end{tabular} &
  \begin{tabular}[c]{@{}l@{}}
  \parbox{5cm}{
\begin{Large}
  \begin{itemize}
  \item no mention of `what' in any related documents.

  \end{itemize}

\end{Large}
  }
  \end{tabular} &
  \begin{tabular}[c]{@{}l@{}}
  \parbox{5cm}{
\begin{Large}
  \begin{itemize}
\item Kyle Allen had options to Stay at the University of Houston for another season, without the promise of ever seeing the football field again.
But after a three turnover performance against Texas Tech on Sept. 23, Allen was benched and replaced by Kyle Postma who took over as the starter.

  \end{itemize}
\end{Large}
  }
  \end{tabular} &
  \begin{tabular}[c]{@{}l@{}}
  \parbox{5cm}{
\begin{Large}
  \begin{itemize}
  \item no mention of `where' in any related documents.

  \end{itemize}
\end{Large}
  }
  \end{tabular} &
  \begin{tabular}[c]{@{}l@{}}
  \parbox{5cm}{
\begin{Large}
  \begin{itemize}
  \item no mention of `why' in any related documents. \\
\end{itemize}
\end{Large}
  }\end{tabular} \\
\bottomrule
\end{tabular}%
}
\caption{Claim: Allen was benched during Houston s game against Texas Tech.}
\label{tab:5WQA_example7}
\end{minipage} \hfill
\begin{minipage}{0.28\linewidth}
\vspace{-8cm}
\centering
\resizebox{\columnwidth}{!}{%
{\tiny
\fbox{%
    \parbox{\columnwidth}{%
    \textcolor{blue}{Allen was benched during Houston s game against Texas Tech.}
    \\
    \textbf{Prphr 1:} Allen was removed from his position while Houston was facing Texas Tech.
     \\
    \textbf{Prphr 2:} Allen was taken off the field during the Houston-Texas Tech match.
     \\
    \textbf{Prphr 3:} Allen was put on the sideline during Houston's contest versus Texas Tech.
     \\
    \textbf{Prphr 4:} Allen was out of the running during Houston's face of against Texas Tech.   
     \\
    \textbf{Prphr 5:} Allen was forbidden from playing during Houston's contest against Texas Tech.  
    }
    }%
    }
}
\vspace{-3mm}
\caption{Claims paraphrased using \texttt{text-davinci-003}}
\label{fig:chatgpt-image7}
\vspace{-3mm}
\end{minipage}
\end{figure*}

\begin{figure*}[!ht]
\begin{minipage}[b]{0.7\linewidth}
\centering
\resizebox{0.98\textwidth}{!}{%
\begin{tabular}{@{}lllll@{}}
\\
\\ \midrule
\begin{LARGE}\textbf{Who claims}\end{LARGE}     & \begin{LARGE}\textbf{What claims}\end{LARGE}    & \begin{LARGE}\textbf{When claims}\end{LARGE}    & \begin{LARGE}\textbf{Where claims}\end{LARGE}   & \begin{LARGE}\textbf{Why claims}\end{LARGE}    
\\
\hline
\begin{tabular}[c]{@{}l@{}}
\parbox{5cm}{
\begin{Large}
  \begin{itemize}
  \item \textbf{Q1}: \textbf{\textit{Who said?}}\\
  \underline{Ans}:  Kamala Harris.\\
  \end{itemize}
\end{Large}
}
\end{tabular} &
  \begin{tabular}[c]{@{}l@{}}
  \parbox{5cm}{
\begin{Large}
  \begin{itemize}
  \item \textbf{Q1}: \textbf{\textit{What did Kamala Harris say?}}\\
  \underline{Ans}:  “if you are going to be standing in that line for all those hours,you can’t have any food."\\
  \end{itemize}
\end{Large}
  }
  \end{tabular} &
  \begin{tabular}[c]{@{}l@{}}
  \parbox{5cm}{
\begin{Large}
    \centering
    No claim!
\end{Large}
  }
   \end{tabular} &
  \begin{tabular}[c]{@{}l@{}}
  \parbox{5cm}{
\begin{Large}
  \begin{itemize}
  \item \textbf{Q1}: \textbf{\textit{Where are people supposed to stand?}}\\
  \underline{Ans}: In line.\\
  \end{itemize}
\end{Large}
  }
  \end{tabular} &
  \begin{tabular}[c]{@{}l@{}}
  \parbox{5cm}{
\begin{Large}
    \centering
    No claim!
\end{Large}
  }
  \end{tabular}\\
  \hline
\includegraphics[width=0.06\textwidth]{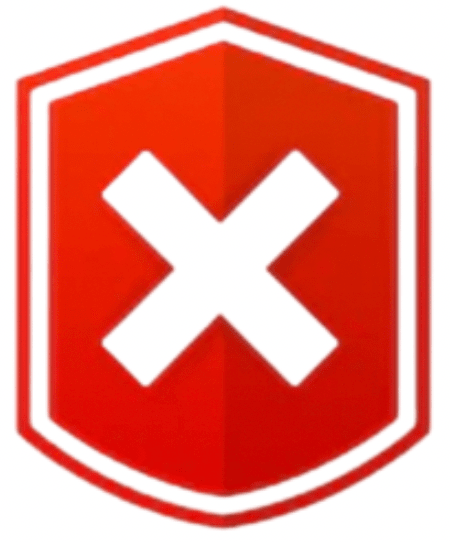} 
\begin{LARGE}\textbf{\textcolor{red}{verified false}}\end{LARGE}
& 
\includegraphics[width=0.06\textwidth]{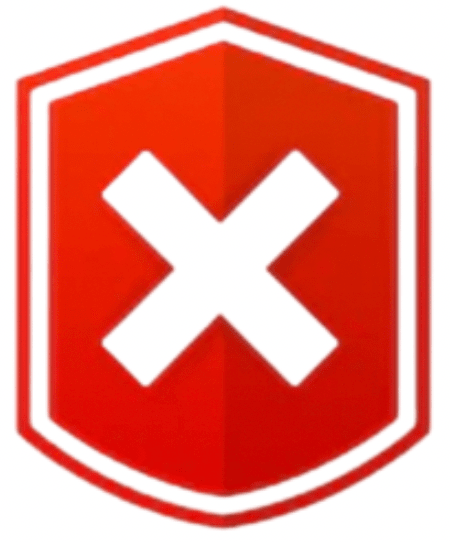} 
\begin{LARGE}\textbf{\textcolor{red}{verified false}}\end{LARGE}
& 
\includegraphics[width=0.04\textwidth]{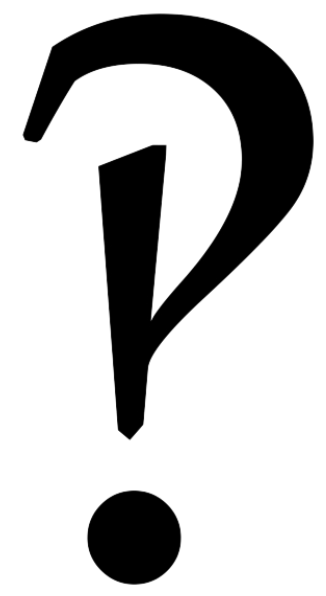} 
\begin{LARGE}\textbf{not verifiable}\end{LARGE}
&
\begin{LARGE}\textbf{\textcolor{green}{verified valid}}\end{LARGE}

& 
\includegraphics[width=0.04\textwidth]{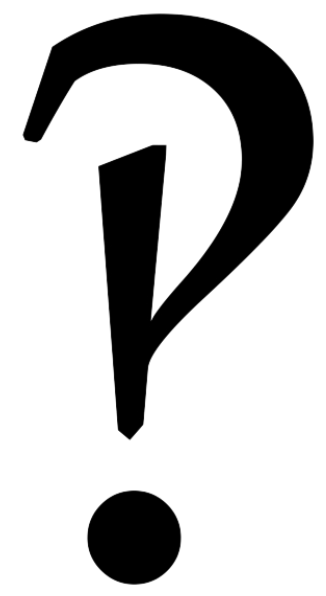} 
\begin{LARGE}\textbf{not verifiable}\end{LARGE}
\\
\hline
\multicolumn{5}{c}{\cellcolor[HTML]{C0C0C0}{
\begin{huge}
\textbf{Evidence}
\end{huge}
}
}  
\\
\hline
\begin{tabular}[c]{@{}l@{}}
\parbox{5cm}{
\begin{Large}
\begin{itemize}
  \item Vice President Kamala Harris said that \textbf{state lawmakers have proposed hundreds of laws that will suppress or make it difficult for people to vote, and that one way state lawmakers have sought to curtail access to ballot is to cut off food or water to voters in line.}
\end{itemize}
\end{Large}
}
\end{tabular} &
  \begin{tabular}[c]{@{}l@{}}
  \parbox{5cm}{
\begin{Large}
  \begin{itemize}
  \item Vice President Kamala Harris said that \textbf{state lawmakers have proposed hundreds of laws that will suppress or make it difficult for people to vote, and that one way state lawmakers have sought to curtail access to ballot is to cut off food or water to voters in line.}

  \end{itemize}

\end{Large}
  }
  \end{tabular} &
  \begin{tabular}[c]{@{}l@{}}
  \parbox{5cm}{
\begin{Large}
  \begin{itemize}
\item no mention of `when' in any related documents. \\

  \end{itemize}
\end{Large}
  }
  \end{tabular} &
  \begin{tabular}[c]{@{}l@{}}
  \parbox{5cm}{
\begin{Large}
  \begin{itemize}
  \item Vice President Kamala Harris said that \textbf{state lawmakers have proposed hundreds of laws that will suppress or make it difficult for people to vote, and that one way state lawmakers have sought to curtail access to ballot is to cut off food or water to voters in line.}

  \end{itemize}
\end{Large}
  }
  \end{tabular} &
  \begin{tabular}[c]{@{}l@{}}
  \parbox{5cm}{
\begin{Large}
  \begin{itemize}
  \item no mention of `why' in any related documents. \\
\end{itemize}
\end{Large}
  }\end{tabular} \\
\bottomrule
\end{tabular}%
}
\caption{Claim: Kamala Harris said that the new and proposed state laws on voting mean "if you are going to be standing in that line for all those hours, you can’t have any food."}
\label{tab:5WQA_example8}
\end{minipage} \hfill
\begin{minipage}{0.28\linewidth}
\vspace{-8cm}
\centering
\resizebox{\columnwidth}{!}{%
{\tiny
\fbox{%
    \parbox{\columnwidth}{%
    \textcolor{blue}{Kamala Harris said that the new and proposed state laws on voting mean "if you are going to be standing in that line for all those hours, you can’t have any food."}
    \\
    \textbf{Prphr 1:} Kamala Harris expressed that the new and intended state regulations on voting mean "in case you are in the queue for all those hours, there is no eatables allowed."
     \\
    \textbf{Prphr 2:} Kamala Harris spoke that the current and planned state legislations related to voting signify "if you are standing in that line for all that time, you cannot have any food."
     \\
    \textbf{Prphr 3:} Kamala Harris highlighted that the recent and put forward state rules on voting mean that there is no food allowed while standing in line.
     \\
    \textbf{Prphr 4:} Kamala Harris has commented on the new state laws on voting, proclaiming that people waiting in the long queue are not able to consume food.
     \\
    \textbf{Prphr 5:} Kamala Harris mentioned that the state regulations being contended for voting have the stipulation that individuals who are standing in line for a prolonged period of time are not allowed to be eating.
    }
    }%
    }
}
\vspace{-3mm}
\caption{Claims paraphrased using \texttt{text-davinci-003}}
\label{fig:chatgpt-image8}
\vspace{-3mm}
\end{minipage}
\end{figure*}

\begin{figure*}[!ht]
\begin{minipage}[b]{0.7\linewidth}
\centering
\resizebox{0.98\textwidth}{!}{%
\begin{tabular}{@{}lllll@{}}
\\
\\ \midrule
\begin{LARGE}\textbf{Who claims}\end{LARGE}     & \begin{LARGE}\textbf{What claims}\end{LARGE}    & \begin{LARGE}\textbf{When claims}\end{LARGE}    & \begin{LARGE}\textbf{Where claims}\end{LARGE}   & \begin{LARGE}\textbf{Why claims}\end{LARGE}    
\\
\hline
\begin{tabular}[c]{@{}l@{}}
\parbox{5cm}{
\begin{Large}
    \centering
    No claim!
\end{Large}
}
\end{tabular} &
  \begin{tabular}[c]{@{}l@{}}
  \parbox{5cm}{
\begin{Large}
  \begin{itemize}
  \item \textbf{Q1}: \textbf{\textit{What begin in the philippines?}}\\
  \underline{Ans}: The start of coronavirus.\\
  \end{itemize}
\end{Large}
  }
  \end{tabular} &
  \begin{tabular}[c]{@{}l@{}}
  \parbox{5cm}{
\begin{Large}
  \begin{itemize}
  \item \textbf{Q1}: \textbf{\textit{when did the woman from wuhan arrive in manila?}}\\
  \underline{Ans}: After traveling to Cebu City.\\
  \end{itemize}
\end{Large}
  }
   \end{tabular} &
  \begin{tabular}[c]{@{}l@{}}
  \parbox{5cm}{
\begin{Large}
    \centering
    No claim!
\end{Large}
  }
  \end{tabular} &
  \begin{tabular}[c]{@{}l@{}}
  \parbox{5cm}{
\begin{Large}
    \centering
    No claim!
\end{Large}
  }
  \end{tabular}\\
  \hline
\begin{LARGE}\textbf{{not verifiable}}\end{LARGE}
& 
\begin{LARGE}\textbf{\textcolor{green}{verified valid}}\end{LARGE}
& 
\begin{LARGE}\textbf{\textcolor{green}{verified valid}}\end{LARGE}
& 
\includegraphics[width=0.04\textwidth]{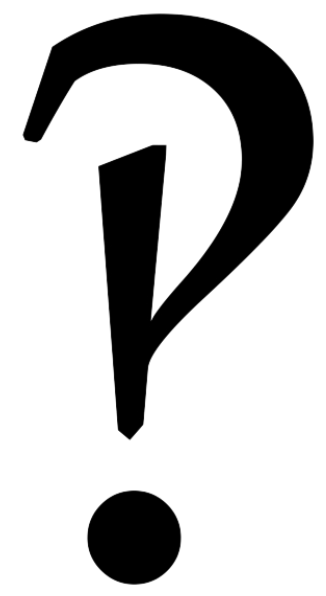} 
\begin{LARGE}\textbf{not verifiable}\end{LARGE}
& 
\includegraphics[width=0.04\textwidth]{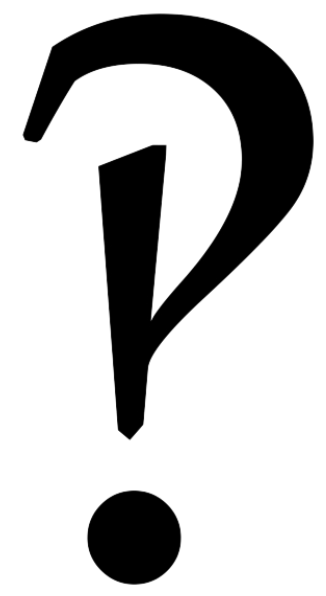} 
\begin{LARGE}\textbf{not verifiable}\end{LARGE}
\\
\hline
\multicolumn{5}{c}{\cellcolor[HTML]{C0C0C0}{
\begin{huge}
\textbf{Evidence}
\end{huge}
}
}  
\\
\hline
\begin{tabular}[c]{@{}l@{}}
\parbox{5cm}{
\begin{Large}
\begin{itemize}
  \item no mention of `who' in any related documents.
\end{itemize}
\end{Large}
}
\end{tabular} &
  \begin{tabular}[c]{@{}l@{}}
  \parbox{5cm}{
\begin{Large}
  \begin{itemize}
  \item Philippine health officials have confirmed the first case of the new coronavirus in the country.A 38-year-old Chinese woman, who arrived in the country from Wuhan, China, on Jan. 21, tested positive for \textbf{the novel coronavirus}, Health Secretary Francisco Duque told a news conference.

  \end{itemize}

\end{Large}
  }
  \end{tabular} &
  \begin{tabular}[c]{@{}l@{}}
  \parbox{5cm}{
\begin{Large}
  \begin{itemize}
\item A 38-year-old Chinese woman, who arrived in the country from Wuhan, China, on Jan. 21, tested positive for the novel coronavirus . DOH Epidemiology Bureau Director Ferchito Avelino said they are also looking at places where the woman stayed in Cebu and Dumaguete. He added that they are working to identify and quarantine employees at establishments who had close contact with the woman.

  \end{itemize}
\end{Large}
  }
  \end{tabular} &
  \begin{tabular}[c]{@{}l@{}}
  \parbox{5cm}{
\begin{Large}
  \begin{itemize}
  \item no mention of `where' in any related documents.

  \end{itemize}
\end{Large}
  }
  \end{tabular} &
  \begin{tabular}[c]{@{}l@{}}
  \parbox{5cm}{
\begin{Large}
  \begin{itemize}
  \item no mention of `why' in any related documents. \\
\end{itemize}
\end{Large}
  }\end{tabular} \\
\bottomrule
\end{tabular}%
}
\caption{Claim: The start of coronavirus in the Philippines was a 38-year-old woman from Wuhan who arrived in Manila after traveling to Cebu City.}
\label{tab:5WQA_example9}
\end{minipage} \hfill
\begin{minipage}{0.28\linewidth}
\vspace{-8cm}
\centering
\resizebox{\columnwidth}{!}{%
{\tiny
\fbox{%
    \parbox{\columnwidth}{%
    \textcolor{blue}{The start of coronavirus in the Philippines was a 38-year-old woman from Wuhan who arrived in Manila after traveling to Cebu City.}
    \\
    \textbf{Prphr 1:} The emergence of coronavirus in the Philippines was sparked by a 38-year-old female from Wuhan who made her way to Manila following her visit to Cebu City.
     \\
    \textbf{Prphr 2:} The onset of coronavirus in the Philippines was initiated by a 38-year-old female from Wuhan who had visited Manila after traveling to Cebu City.   
     \\
    \textbf{Prphr 3:} The beginning of coronavirus in the Philippines was started by a 38-year-old female from Wuhan who moved to Manila after going to Cebu City.
     \\
    \textbf{Prphr 4:} The initial appearance of the coronavirus in the Philippines came from a 38-year-old female from Wuhan who journeyed to Manila via Cebu City.
     \\
    \textbf{Prphr 5:} The first time the coronavirus arrived in the Philippines was with a 38-year-old female from Wuhan that stopped by Manila after a trip to Cebu City.
    }
    }%
    }
}
\vspace{-3mm}
\caption{Claims paraphrased using \texttt{text-davinci-003}}
\label{fig:chatgpt-image9}
\vspace{-3mm}
\end{minipage}
\end{figure*}

\begin{figure*}[!ht]
\begin{minipage}[b]{0.7\linewidth}
\centering
\resizebox{0.98\textwidth}{!}{%
\begin{tabular}{@{}lllll@{}}
\\
\\ \midrule
\begin{LARGE}\textbf{Who claims}\end{LARGE}     & \begin{LARGE}\textbf{What claims}\end{LARGE}    & \begin{LARGE}\textbf{When claims}\end{LARGE}    & \begin{LARGE}\textbf{Where claims}\end{LARGE}   & \begin{LARGE}\textbf{Why claims}\end{LARGE}    
\\
\hline
\begin{tabular}[c]{@{}l@{}}
\parbox{5cm}{
\begin{Large}
  \begin{itemize}
  \item \textbf{Q1}: \textbf{\textit{Who wrote the book series that robbie coltrane is based on?}}\\
  \underline{Ans}: By J. K. Rowling.\\
  \end{itemize}
\end{Large}
}
\end{tabular} &
  \begin{tabular}[c]{@{}l@{}}
  \parbox{5cm}{
\begin{Large}
  \begin{itemize}
  \item \textbf{Q1}: \textbf{\textit{What is robbie coltrane known for?}}\\
  \underline{Ans}: For his roles as a fictional character.\\
  \end{itemize}
\end{Large}
  }
  \end{tabular} &
  \begin{tabular}[c]{@{}l@{}}
  \parbox{5cm}{
\begin{Large}
    \centering
    No claim!
\end{Large}
  }
   \end{tabular} &
  \begin{tabular}[c]{@{}l@{}}
  \parbox{5cm}{
\begin{Large}
    \centering
    No claim!
\end{Large}
  }
  \end{tabular} &
  \begin{tabular}[c]{@{}l@{}}
  \parbox{5cm}{
\begin{Large}
    \centering
    No claim!
\end{Large}
  }
  \end{tabular}\\
  \hline
\begin{LARGE}\textbf{\textcolor{green}{verified valid}}\end{LARGE}
& 
\begin{LARGE}\textbf{\textcolor{green}{verified valid}}\end{LARGE}
& 
\includegraphics[width=0.06\textwidth]{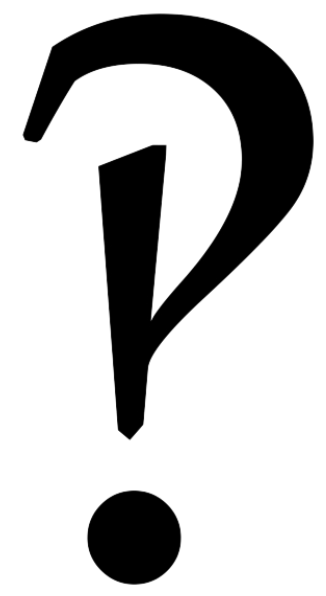} 
\begin{LARGE}\textbf{{not verifiable}}\end{LARGE}
& 
\includegraphics[width=0.04\textwidth]{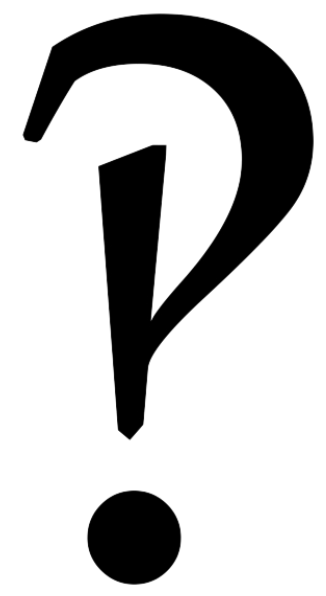} 
\begin{LARGE}\textbf{not verifiable}\end{LARGE}
& 
\includegraphics[width=0.04\textwidth]{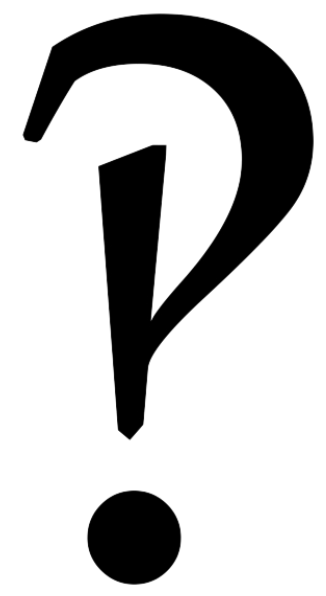} 
\begin{LARGE}\textbf{not verifiable}\end{LARGE}
\\
\hline
\multicolumn{5}{c}{\cellcolor[HTML]{C0C0C0}{
\begin{huge}
\textbf{Evidence}
\end{huge}
}
}  
\\
\hline
\begin{tabular}[c]{@{}l@{}}
\parbox{5cm}{
\begin{Large}
\begin{itemize}
  \item Coltrane was widely known for starring in the "Harry Potter" franchise, based on the books by \textbf{J.K. Rowling}, alongside Daniel Radcliffe in the title role. 
\end{itemize}
\end{Large}
}
\end{tabular} &
  \begin{tabular}[c]{@{}l@{}}
  \parbox{5cm}{
\begin{Large}
  \begin{itemize}
  \item Anthony Robert McMillan OBE (30 March 1950 – 14 October 2022), known professionally as Robbie Coltrane, was a Scottish actor and comedian. \textbf{He gained worldwide recognition in the 2000s for playing Rubeus Hagrid in the Harry Potter film series}.

  \end{itemize}

\end{Large}
  }
  \end{tabular} &
  \begin{tabular}[c]{@{}l@{}}
  \parbox{5cm}{
\begin{Large}
  \begin{itemize}
  \item no mention of `when' in any related documents. 

  \end{itemize}
\end{Large}
  }
  \end{tabular} &
  \begin{tabular}[c]{@{}l@{}}
  \parbox{5cm}{
\begin{Large}
  \begin{itemize}
  \item no mention of `where' in any related documents.

  \end{itemize}
\end{Large}
  }
  \end{tabular} &
  \begin{tabular}[c]{@{}l@{}}
  \parbox{5cm}{
\begin{Large}
  \begin{itemize}
  \item no mention of `why' in any related documents. \\
\end{itemize}
\end{Large}
  }\end{tabular} \\
\bottomrule
\end{tabular}%
}
\caption{Claim: Robbie Coltrane is known for his film roles as a fictional character based on a book series written by J. K. Rowling.}
\label{tab:5WQA_example10}
\end{minipage} \hfill
\begin{minipage}{0.28\linewidth}
\vspace{-8cm}
\centering
\resizebox{\columnwidth}{!}{%
{\tiny
\fbox{%
    \parbox{\columnwidth}{%
    \textcolor{blue}{Robbie Coltrane is known for his film roles as a fictional character based on a book series written by J. K. Rowling.}
    \\
    \textbf{Prphr 1:} Robbie Coltrane is famed for his movie performances of a fictional character inspired by a set of books written by J. K. Rowling.
     \\
    \textbf{Prphr 2:} Robbie Coltrane is renowned for his film parts as a fictional character originated from a book series composed by J. K. Rowling.
     \\
    \textbf{Prphr 3:} Robbie Coltrane is well-known for his parts in films inspired by a book collection from J. K. Rowling.
     \\
    \textbf{Prphr 4:} Robbie Coltrane rose to popularity because of the parts he played in films based off of the fictional work of J. K. Rowling.   
     \\
    \textbf{Prphr 5:} Robbie Coltrane is admired for his roles in pictures as a fictional character drawn from a collection of literature written by J. K. Rowling. 
    }
    }%
    }
}
\vspace{-3mm}
\caption{Claims paraphrased using \texttt{text-davinci-003}}
\label{fig:chatgpt-image10}
\vspace{-3mm}
\end{minipage}
\end{figure*}

\end{document}